\PassOptionsToPackage{unicode}{hyperref}
\PassOptionsToPackage{hyphens}{url}
\documentclass[
  10pt,
  letterpaper,
]{article}
\usepackage{amsmath,amssymb}
\usepackage{iftex}
\ifPDFTeX
  \usepackage[T1]{fontenc}
  \usepackage[utf8]{inputenc}
  \usepackage{textcomp} 
\else 
  \usepackage{unicode-math} 
  \defaultfontfeatures{Scale=MatchLowercase}
  \defaultfontfeatures[\rmfamily]{Ligatures=TeX,Scale=1}
\fi
\usepackage{lmodern}
\ifPDFTeX\else
\fi
\IfFileExists{upquote.sty}{\usepackage{upquote}}{}
\IfFileExists{microtype.sty}{
  \usepackage[]{microtype}
  \UseMicrotypeSet[protrusion]{basicmath} 
}{}
\makeatletter
\@ifundefined{KOMAClassName}{
  \IfFileExists{parskip.sty}{%
    \usepackage{parskip}
  }{
    \setlength{\parindent}{0pt}
    \setlength{\parskip}{6pt plus 2pt minus 1pt}}
}{
  \KOMAoptions{parskip=half}}
\makeatother
\usepackage{xcolor}
\usepackage{color}
\usepackage{fancyvrb}

\DefineVerbatimEnvironment{Highlighting}{Verbatim}{commandchars=\\\{\}}
\newenvironment{Shaded}{}{}

\newcommand{\NormalTok}[1]{#1}

\usepackage{longtable,booktabs,array}
\usepackage{calc} 
\usepackage{etoolbox}
\makeatletter
\patchcmd\longtable{\par}{\if@noskipsec\mbox{}\fi\par}{}{}
\makeatother
\IfFileExists{footnotehyper.sty}{\usepackage{footnotehyper}}{\usepackage{footnote}}
\makesavenoteenv{longtable}
\usepackage{graphicx}
\makeatletter
\def\maxwidth{\ifdim\Gin@nat@width>\linewidth\linewidth\else\Gin@nat@width\fi}
\def\maxheight{\ifdim\Gin@nat@height>\textheight\textheight\else\Gin@nat@height\fi}
\makeatother
\setkeys{Gin}{width=\maxwidth,height=\maxheight,keepaspectratio}
\makeatletter
\def\fps@figure{htbp}
\makeatother
\providecommand{\tightlist}{%
  \setlength{\itemsep}{0pt}\setlength{\parskip}{0pt}}
\ifLuaTeX
\usepackage[bidi=basic]{babel}
\else
\usepackage[bidi=default]{babel}
\fi
\babelprovide[main,import]{american}
\ifPDFTeX
\else
\babelfont{rm}[BoldFont=texgyretermes-bold.otf,ItalicFont=texgyretermes-italic.otf,BoldItalicFont=texgyretermes-bolditalic.otf]{texgyretermes-regular.otf}
\fi

\def\languageshorthands#1{}
\makeatletter
\def\input@path{{tex/}}
\makeatother
\usepackage[preprint,nonatbib]{neurips_2026}

\providecommand{\doi}[1]{URL \url{https://doi.org/#1}}

\allowdisplaybreaks
\numberwithin{equation}{section}
\ifLuaTeX
  \usepackage{selnolig}  
\fi
\usepackage[]{natbib}
\IfFileExists{bookmark.sty}{\usepackage{bookmark}}{\usepackage{hyperref}}
\IfFileExists{xurl.sty}{\usepackage{xurl}}{} 
\hypersetup{
  pdftitle={Dalek: A Constructive Agent Machine},
  pdfauthor={Wanpeng Xie wanpeng.xie@gmail.com},
  pdflang={en-US},
  pdfkeywords={agent systems, self-reproducing automata, self-modifying
systems, constructive definition, software architecture},
  hidelinks,
  pdfcreator={LaTeX via pandoc}}

\title{Dalek: A Constructive Agent Machine}
\usepackage{etoolbox}
\makeatletter
\providecommand{\subtitle}[1]{
  \apptocmd{\@title}{\par {\large #1 \par}}{}{}
}
\makeatother
\subtitle{Self-Maintenance, Self-Evolution, Self-Reproduction, and
Self-Organization by Construction}
\author{Wanpeng Xie\\
\texttt{wanpeng.xie@gmail.com}}
\date{}

\begin{document}
\maketitle
\begin{abstract}
We present Dalek, a closed machine designed for agents that realizes
self-maintenance, self-evolution, self-reproduction, and
self-organization on any substrate satisfying a general host contract.
The machine is built from three primitives---actors, messages, and
channels. Four obligations---a host boundary, a construction language,
admissible transitions, and rule heredity---give its boundary, identity,
and closure a structural basis.

Von Neumann's 1948 self-reproducing automaton supplies a hereditary
constructional core: a self-description together with a constructor, a
copier, and a controller. Dalek combines this core with the four
obligations and rederives its medium for a text-and-message agent
substrate, adding explicit structures for boundary, identity, history,
and growth. A large language model and a compiler occupy the payload
position and form a general capability producer. New capabilities are
authored, compiled, installed into the description, and inherited by
descendants. The same path produces the machine's own organs and even
its runtime, closing heredity and evolution within the machine.

\textbf{Keywords:} agent systems; self-reproducing automata;
self-modifying systems; constructive definition; software architecture
\end{abstract}

\newpage

{
\setcounter{tocdepth}{1}
\tableofcontents
}
\hypertarget{introduction}{%
\section{Introduction}\label{introduction}}

\hypertarget{from-an-inventory-to-a-machine}{%
\subsection{From an Inventory to a
Machine}\label{from-an-inventory-to-a-machine}}

Agents are commonly implemented as collections of components around a
large language model. The model generates; memory, tools, workflows,
subagents, gateways, and validators do the rest. One response to a
harder task is to lengthen this inventory, assigning a new kind of
component to each new kind of demand. As the inventory grows, so do the
costs of assembly, configuration, and maintenance. Capability is
purchased with system complexity.

A second response is to let the model generate and assemble components
at runtime. Dynamic plug-ins and code mode are representative examples
\citep{Cordis26, deepseek2026harness}. Generation replaces
preinstallation, so new capabilities no longer have to be enumerated in
advance. The deeper problem begins after generation. Code written by an
agent may be callable immediately, yet whether that code has become part
of the agent, survives a restart, is inherited by a copy, or is
authorized for installation remains a decision of the file system,
loader, image builder, deployment script, or operator. Whether the same
code is merely the product of a task or a modification of the system
itself depends on arrangements outside the agent. The ability to
generate new structure has entered the agent; the authority to decide
what constitutes the agent has not.

Uncontrolled complexity and displaced constitutive authority have the
same source: \textbf{the absence of a constructive definition of the
agent machine}.

This absence is immaterial when an agent serves a predetermined task and
remains under continuous human maintenance: a tool need not define
itself. The requirement changes in kind when tasks cannot be enumerated
in advance, operation does not presuppose a human in attendance, and the
object of improvement eventually includes the system's own structure.
The old claim was existential: the system completed a task, ran
unattended for several hours, or once rewrote its own scaffold. A single
demonstration could establish it. The new claim is invariant: however
long the system runs, whatever tasks arrive, and however often its
structure changes, the properties in §1.2 continue to hold. A
demonstration can establish that something happened once. To say that it
remains true requires a structural guarantee; before such a guarantee
can be stated, the machine that bears it must be defined.

\hypertarget{criteria-for-a-constructive-definition}{%
\subsection{Criteria for a Constructive
Definition}\label{criteria-for-a-constructive-definition}}

We study a particular class of objects: agent machines that can
incorporate capabilities produced by open-ended tasks while supporting
claims about long-running operation and structural change. Relative to a
host contract \(\Omega\), such a machine has four definitional
obligations:

\begin{enumerate}
\def\labelenumi{\arabic{enumi}.}
\tightlist
\item
  \textbf{Host boundary.} Every mechanism that can affect a claimed
  property either belongs to the machine or is explicitly assigned to
  \(\Omega\) as an input or host assumption. There is no undeclared
  third location. The boundary determines both the scope of a claim and
  the locus to which a property is attributed.
\item
  \textbf{Construction language.} Legal machine forms are generated from
  finite primitives by finite composition rules. New capabilities may be
  produced without a predetermined bound; the same language determines
  when they become part of the machine and in what form they persist.
\item
  \textbf{Admissible transitions.} The initial machine and the
  operations that alter its constitution are specified explicitly. Every
  admissible transition maps a legal machine back into the same machine
  class. Legal states and admissible transitions together form the
  machine's operational semantics.
\item
  \textbf{Rule heredity.} If self-improvement is to be attributed to the
  machine, the rules that produce a legal successor must themselves be
  representable, constructible, and heritable. The current rules remain
  fixed during one construction; revised rules are written into the
  successor. This temporal separation avoids the vicious circle of
  constructing oneself at the same instant.
\end{enumerate}

The first three obligations define the machine's carrier and dynamics.
The fourth places the rules that generate those dynamics inside the
generational relation. A system whose loader or construction service is
explicitly assigned to \(\Omega\) remains well-defined under the first
three obligations; its construction capability then belongs to the host,
and the fourth obligation does not hold.

These obligations define a machine class rather than a unique
architecture. Any system that satisfies them is a member of that class.
This paper establishes existence: Dalek is one complete running witness.

\hypertarget{necessity-and-the-governance-surface}{%
\subsection{Necessity and the Governance
Surface}\label{necessity-and-the-governance-surface}}

All four obligations arise from the proof form required by an unbounded
claim. Let \(P\) mean either ``this is still a legal machine'' or ``this
machine still carries the mechanisms required to construct a
successor.'' To establish \(P\) over an arbitrarily long legal history,
a finite argument supplies a base case and proves that each admissible
operation maps a legal machine back into the same class. The object of
induction needs a definite domain: the host boundary and construction
language. The induction step must compose indefinitely: admissible
transitions. When the rules that form the induction step enter the
domain of change, rule heredity is required as well.

Open-ended tasks alone require only generative finite means: a fixed
universal interpreter can already process programs that were not
enumerated in advance. Construction language and admissible transitions
become obligations of the machine when generated capabilities must also
be installed, persisted, restarted, and inherited---when the task axis
meets the self-modification axis. A system whose rules remain external
can likewise be well-defined by declaring those rules part of
\(\Omega\). Rule heredity becomes necessary only when improvement is
attributed to the machine itself. The result is a set of obligations on
a machine class, not a unique implementation.

A constitutive boundary also yields a two-sided benefit. If all
organizational effects between machine and environment pass through a
finite interface, monitoring and intervention need cover only a finite
governance surface. The environment cannot rewrite the machine outside
its admissible transitions, while the machine's organizational effects
acquire definite points of authorization, audit, revocation, and
termination. Structural change either preserves these mediation points
or records their alteration as an explicit transition, allowing
governance to persist across generations.

Our claims concern organizational effects: members, messages, doors, and
changes in constitution belong to the machine. Direct file or network
access performed by an actor during one invocation is not part of the
machine's history (§3.2.3). Extending the same criteria to every
physical effect would require a stronger \(\Omega\).

\hypertarget{dalek-construction-and-witness}{%
\subsection{Dalek: Construction and
Witness}\label{dalek-construction-and-witness}}

Dalek is constructed from three sources. Von Neumann's 1948
self-reproducing automaton supplies a hereditary constructional core
\citep{vN48}. The four obligations require an explicit boundary, a
finite construction language, admissible constitutive transitions, and
rules that enter heredity. The properties of an agent substrate make a
uniform actor/message interface and channel boundaries the natural
medium. Their combination yields a new agent machine rather than a new
implementation of von Neumann's machine. \(\Omega\) supplies only
execution, storage, and networking. Actors collaborate through messages;
channels provide organizational boundaries. \(G\) describes what the
machine ought to be, while \(H\) records what it has undergone. A
runtime \(R\) that is blind to function and organization defines the
constitutive transitions. The constructor \(A\), copier \(B\),
controller \(C\), and capability producer \(D\) are themselves contained
in \(G\), so the present generation can construct a successor carrying
revised organs and even a revised runtime.

This construction moves the organizing center of an agent system from a
task loop to the lifecycle of an individual. Installation, restart,
reproduction, and upgrade cease to be maintenance operations performed
around the agent and become transitions of the machine itself. Task
capabilities become members that can grow, combine, and reproduce within
that lifecycle. The four \emph{selves} name four directions of the same
lifecycle: maintenance, change, continuation, and association.

The paper makes three contributions. First, it states the constructive
obligations of an agent machine (§§1.2--1.3). Second, it constructs
Dalek, in which all four obligations are jointly realized (§§2--3).
Third, it provides ledger evidence that the same structure delivers
self-maintenance, self-evolution, self-reproduction, and
self-organization (§4).

The rest of the paper proceeds as follows. Section 2 states the
inherited construction and its assumptions, then selects a substrate.
Section 3 builds the machine. Section 4 lets its ledgers provide
line-by-line evidence and defines the semantics of failed paths and the
boundary of liveness. Section 5 discusses the consequences and scope of
the definition. Section 6 reviews related work. Section 7 concludes and
identifies conservative extensions.

\hypertarget{constructional-model-and-substrate}{%
\section{Constructional Model and
Substrate}\label{constructional-model-and-substrate}}

Of the four definitional obligations in §1.2, rule heredity already has
a constructional core: von Neumann's 1948 self-reproducing automaton
\citep{vN48}.\footnote{``Von Neumann architecture'' commonly denotes the
  1945 stored-program organization of arithmetic, control, memory,
  input, and output \citep{vN45}. This paper instead uses von Neumann's
  other construction, presented in 1948: the self-reproducing automaton.
  The former organizes a computation; the latter supplies a
  constructional model for a machine capable of constructing and copying
  itself. Dalek uses the second as one core in the construction of an
  agent machine.} We retain this core (§2.1), combine it with the
remaining obligations, and derive a machine on an agent-specific
substrate (§2.2). This section defines the notation \(A\), \(B\), \(C\),
\(D\), and \(G\), together with quasi-quiescence and the host contract.

\hypertarget{the-inherited-construction}{%
\subsection{The Inherited
Construction}\label{the-inherited-construction}}

The construction consists of four components and one description
\citep{vN48}. \textbf{\(A\) is a universal constructor}: given a
description \(G\), it constructs the automaton described by \(G\). \(A\)
remains fixed; complexity resides in \(G\), which grows with the
complexity of the object to be built. \textbf{\(B\) is a copier}: it
copies \(G\) without interpreting it. \textbf{\(C\) is a controller}: it
first directs \(A\) to construct from \(G\), then directs \(B\) to copy
\(G\) and insert the copy into the new construction, after which it
detaches and starts the offspring. \textbf{\(D\) is a payload}: an
arbitrary automaton carried by the machine but unused by the logic of
construction and copying. The complete machine is

\[
E = (A + B + C + D) + G,
\]

where \(G\) describes the entire assembly \(A+B+C+D\). One cycle of
\(E\) produces a verbatim copy of \(E\).\footnote{\textbf{Notation.} In
  von Neumann's notation, the description is \(I\) (or \(I_{D+F}\) when
  a payload is present), \(D\) denotes the assembly \(A+B+C\), \(F\) is
  the additional payload, and the complete machine is \(E_F=D+I_{D+F}\).
  Dalek retains the meanings of \(A\), \(B\), and \(C\), uses \(D\) for
  the payload, and writes the description as \(G\). Von Neumann's \(I\)
  and Dalek's \(G\) name the same architectural role in different media.
  Section 3.4.7 lists every notational and structural departure.} The
same \(G\) is used twice: \(A\) interprets it and \(B\) copies it. \(G\)
does not contain a description of itself; copying regenerates it, so the
body and its description are formed in a definite temporal order
\citep{vN48}.

\begin{figure}
\centering
\includegraphics[width=1\textwidth,height=\textheight]{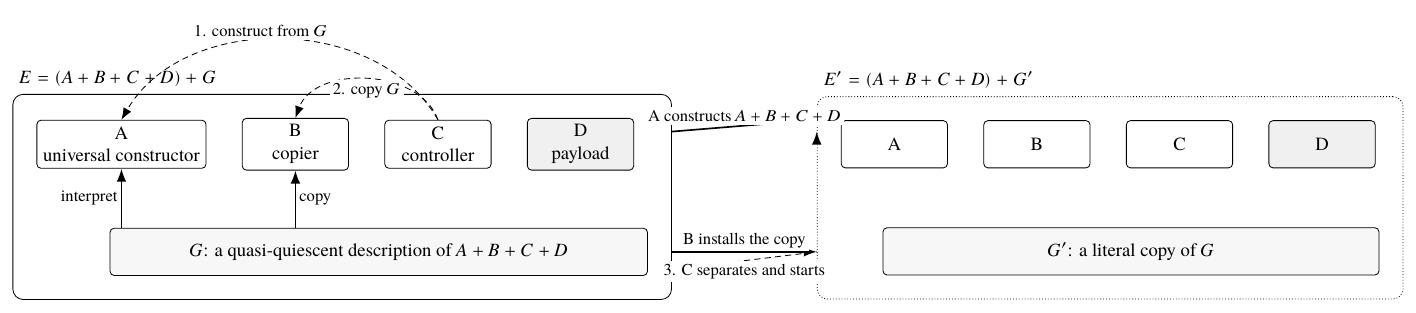}
\caption{Von Neumann's 1948 construction: \(E=(A+B+C+D)+G\). The same
\(G\) is used twice---interpreted by \(A\) and copied by \(B\). Under
the direction of \(C\), \(A\) constructs \(A+B+C+D\), \(B\) inserts a
copy of \(G\), and the resulting machine is detached and started. \(D\)
is a payload unused by construction and copying.}
\end{figure}

The construction relies on two assumptions. First, the description is
\textbf{quasi-quiescent}: it remains unchanged during copying. A live
original reacts and changes as it is examined and therefore cannot be
copied piece by piece; a linear chain that remains fixed for the
duration of copying can \citep{TSRA}. Second, physics is supplied by the
environment. The construction is kinematic: the machine is immersed in a
sea of parts; components are axiomatized as black boxes by specifying
only stimulus and response; energy and motion belong to the environment
\citep{vN48}. Section 2.2 gives an implementation of this
\textbf{environment contract}. The substrate chosen here satisfies both
assumptions.

Constructional model and substrate are distinct. The 1948 construction
is stated in terms of black-box components and an environment contract
and does not depend on a particular component physics. The 29-state
cellular automaton devised in 1952--53 is one later realization
\citep{TSRA}, not the construction itself. At the level of the
hereditary core, Dalek preserves the 1948 construction while replacing
its substrate. The complete Dalek architecture also adds the structures
required for boundary, identity, history, and growth.

Measured against the four obligations, the 1948 model supplies the core
of rule heredity and the skeleton of a construction language. Three
things are absent. \textbf{Identity:} the model cannot answer ``which
individual am I?'' \citep{McM00}; §3.4.6 supplies an answer.
\textbf{History and growth:} it produces copies but retains no history,
and actual growth of complexity remained an open problem in the
subsequent literature \citep{McM00}. Dalek records history in a ledger
(§3.2.3), while an internal loop between author and validator admits
growth into the description (§3.4.3). \textbf{A live author:} the
model's payload is inactive in construction and copying, and mutations
confined to that payload are nonlethal \citep{vN48}. Dalek places a
component that can write in the same position and retains the notation
\(D\). The medium, laws, and identity around that component are
developed in §3.

\hypertarget{from-a-constructional-model-to-an-agent-machine}{%
\subsection{From a Constructional Model to an Agent
Machine}\label{from-a-constructional-model-to-an-agent-machine}}

The 1948 model specifies causal roles and relations, not a software
implementation. Its construction requires five things: assemblable
parts, a quiescent and copyable description, organs formed from groups
of parts, a detachable entity, and an environment that supplies the
physics. It also leaves a payload position outside the logic of
reproduction.

Dalek combines this hereditary core with the four obligations of §1.2
and with the properties of an agent-specific substrate. The obligations
require a declared host boundary, a finite language of construction and
change, and rules that enter heredity. The substrate presents
heterogeneous participants through text and message exchange. Together
they yield the following organization. These correspondences are results
of constructing an agent machine, not a one-to-one implementation of von
Neumann's machine.

\begin{enumerate}
\def\labelenumi{\arabic{enumi}.}
\tightlist
\item
  \textbf{Part = actor.} An actor is a behavioral component that
  interacts only through messages \citep{Hew73}. Every constitutive
  member is presented through the same message interface; programs,
  models, tools, humans, and external systems are adapted as actors. The
  constructional benefit is uniformity: there is one representation of a
  part, so \(A\) needs only one construction path.
\item
  \textbf{Description = \(G\).} In this medium, \(G\) is text: static,
  manipulable, and packageable constitutive data. Once written, a
  blueprint is itself a part of the medium. Interpretation by \(A\) and
  copying by \(B\) require no translation across physical kinds, and
  quasi-quiescence follows directly.
\item
  \textbf{Organ boundary = channel.} A channel encloses a group of
  actors in a local boundary and maintains a ledger for them. A channel
  does not itself execute.\footnote{The organs---the constructor \(A\),
    copier \(B\), controller \(C\), and general capability producer
    \(D\)---are organized at channel granularity; execution is carried
    by actors. For engineering convenience, the three functions \(A\),
    \(B\), and \(C\) reside in one initial channel. Together they form
    the smallest organ capable of performing an initial construction
    (§3.4.1).}
\item
  \textbf{Detachable entity = Space.} A message medium has no intrinsic
  space, so the entity denoted by ``detach and start'' must be supplied
  by the construction. A Space comprises a set of channels and one
  runtime. It is the complete machine that \(\Omega\) can start;
  detachment and startup operate on a Space (§3).
\item
  \textbf{Environment = \(\Omega + R\).} Dalek factors the
  undifferentiated physics of the 1948 model into two layers. The
  external host contract \(\Omega\) contains three primitive
  capabilities---execution, storage, and networking---while the internal
  runtime \(R\) enforces the laws of the medium and is inherited with
  the machine. Together, \(\Omega+R\) instantiate the environment
  contract.
\item
  \textbf{Payload position (Dalek's additional requirement) = \(D\), a
  general capability producer.} \(D\) consists of a large language model
  \textbf{\(L\)} and a compiler \textbf{\(U\)}. \(L\) produces candidate
  capabilities. It receives and emits text, its output is probabilistic,
  it retains no trustworthy state across invocations, and its interior
  remains a black box. Its native operational form is therefore one
  incoming message followed by zero or more outgoing messages. \(U\) is
  separate from the author: it compiles candidates and admits those that
  pass into the machine.
\end{enumerate}

The resulting architecture therefore has three distinct inputs. The 1948
model contributes the two uses of \(G\), the division into constructor,
copier, controller, and payload, their temporal order, and
construction-time quasi-quiescence. The definitional obligations,
derived from the agent problem, require the explicit boundary,
construction language, constitutive semantics, and rules of succession.
Preserving a changing individual over time further requires identity and
history, while open-ended tasks require a path for actual growth. The
physics of the available components selects the medium. Computer
organization follows component physics \citep{vN45}; so does this
machine. Everything \(L\) can perceive or do appears as text flowing in
and out. Humans, environments, tools, and other agents therefore meet
\(L\) under one identity---actor---and their interactions take one
form---message. The adaptation layer is localized inside an ordinary
actor program: \(L\)'s loop assembles a ledger view, calls a remote
model, parses its output, and emits individual calls. The medium assigns
no special semantics to \(L\).

The resulting fit can be stated directly:

\begin{longtable}[]{@{}
  >{\raggedright\arraybackslash}p{(\columnwidth - 2\tabcolsep) * \real{0.5000}}
  >{\raggedright\arraybackslash}p{(\columnwidth - 2\tabcolsep) * \real{0.5000}}@{}}
\toprule\noalign{}
\begin{minipage}[b]{\linewidth}\raggedright
Dalek component or property
\end{minipage} & \begin{minipage}[b]{\linewidth}\raggedright
Why it fits \(L\)
\end{minipage} \\
\midrule\noalign{}
\endhead
\bottomrule\noalign{}
\endlastfoot
message / call: one constitutive action & Text is the native input and
output form of both \(L\) and a human participant. \\
\(G\) (description): dual use of description and rule heredity & \(L\)
can read and write static text directly. \\
\(H\) (ledger): identity, recovery, and factual witness & The
probabilistic output of \(L\) cannot be recovered by recomputation;
history preserves it. \\
\(U\) (compiler): separation of variation and selection & \(L\) should
not be both author and judge. \\
no privileged actor & \(L\) can be replaced or coexist with another
\(L\). \\
\end{longtable}

\hypertarget{the-machine}{%
\section{The Machine}\label{the-machine}}

This section constructs Dalek from the three sources established in
§2.2: the hereditary core of the 1948 model, the definitional
obligations of §1.2, and the properties of an agent-specific substrate.
Sections 3.1 and 3.3 define \(\Omega\) and \(R\); §3.2 defines actors,
messages, channels, Spaces, and ledgers; §3.4 defines the
interpretation, copying, and heredity of \(G\). Once the machine has
been assembled, §3.5 checks it against the four obligations.

\hypertarget{the-host-contract-omega}{%
\subsection{\texorpdfstring{The Host Contract
\(\Omega\)}{The Host Contract \textbackslash Omega}}\label{the-host-contract-omega}}

\(\Omega\) is the external half of the environment contract (§2.2): a
fixed, finite capability list containing none of the vocabulary
introduced by this paper. A machine is software running on \(\Omega\);
\(\Omega\) knows nothing of ledgers, organs, Spaces, or reproduction.
The internal half is the runtime \(R\) (§3.3).

\hypertarget{three-capabilities}{%
\subsubsection{Three Capabilities}\label{three-capabilities}}

\begin{longtable}[]{@{}
  >{\raggedright\arraybackslash}p{(\columnwidth - 4\tabcolsep) * \real{0.3333}}
  >{\raggedright\arraybackslash}p{(\columnwidth - 4\tabcolsep) * \real{0.3333}}
  >{\raggedright\arraybackslash}p{(\columnwidth - 4\tabcolsep) * \real{0.3333}}@{}}
\toprule\noalign{}
\begin{minipage}[b]{\linewidth}\raggedright
Capability
\end{minipage} & \begin{minipage}[b]{\linewidth}\raggedright
Contract
\end{minipage} & \begin{minipage}[b]{\linewidth}\raggedright
Experimental realization
\end{minipage} \\
\midrule\noalign{}
\endhead
\bottomrule\noalign{}
\endlastfoot
Execution (\textbf{Exec}) & \texttt{load(source,\ environment)} \(\to\)
callable object; \texttt{spawn(source,\ arguments)} \(\to\) process &
Python 3: in-process \texttt{exec} and child processes \\
Storage (\textbf{Store}) &
\texttt{read\ /\ write\ /\ append(path,\ bytes)}; append is atomic and
durable & File system \\
Network (\textbf{Port}) & \texttt{send(endpoint,\ bytes)};
\texttt{recv(endpoint)} \(\to\) sequence of received bytes & File
inboxes \\
\end{longtable}

These are three semantic roles. Together they form a small sufficient
contract; in the first implementation, the receiving side of Port is
implemented by appending to files in Store. \textbf{Execution includes
processes, not merely evaluation.} Source must become an independently
living process that can be stopped. Without a separate process, an
offspring cannot outlive its parent and self-reproduction degenerates
into a function call. \textbf{Networking includes reception, not merely
transmission.} A conversation with a human, a description sent by
another machine, and a line written by a parent through an offspring's
root door all require a receiving endpoint on the machine side. In the
single-host experiments, that endpoint is a file inbox; across hosts it
is a network endpoint. \textbf{Storage includes atomic append.} The
first law of the medium---append only (§3.2.3)---rests on this
operation.

\hypertarget{boundary}{%
\subsubsection{Boundary}\label{boundary}}

\(\Omega\) supplies no permissions, scheduling or message semantics,
replay or validation, or isolation. These belong, respectively, to the
machine laws (§3.3.4), the medium (§3.2), the ledger (§3.2.3), and
directory conventions. The host boundary of §1.2 now has a concrete
form: none of this paper's terms appears in the interface table of
\(\Omega\). Every mechanism affecting a claimed property is either
inside the machine or on that table. Any underlying system satisfying
the three capabilities in §3.1.1 is a legal \(\Omega\), and machine
identity does not depend on a particular host. The first \(\Omega\) used
in the experiments is Linux, Python 3, and a file system. Section 3.3.5
identifies the layer that changes inside a machine when the host is
changed.

\hypertarget{exogenous-input-and-time}{%
\subsubsection{Exogenous Input and
Time}\label{exogenous-input-and-time}}

The interface table contains neither a clock nor a random-number source.
Any value that crosses the machine's organizational boundary into the
medium and cannot be recomputed from existing state and a deterministic
program is exogenous input and must enter a ledger (§3.2.3). Human
messages, bytes arriving through a door, member replies, time, failures,
and concurrent order all cross the same point: the call boundary. Values
obtained directly from the host during an actor invocation---a remote
response received by \(L\) or a file read---become machine facts only
when the actor emits them as a call, a reply, or an incoming message.
Every step can therefore either be recomputed and compared or copied
from the ledger.

Time follows the same rule. Quiescence in a running machine is the
operational counterpart of quasi-quiescence (§2.1): when there is no
message, nothing moves. A clock inside the body manufactures events and
turns quiescence from a property into an accident. The machine does not
produce time; it organizes time. A machine that requires ticks
subscribes to an external tick source. Each tick enters through a door
as an ordinary stamped message. Physical time thereby leaves the
machine, whose interior retains only the causal order of events
\citep{Lam78}. Runtime invocations are run to completion, one event per
invocation (§3.3.1), and need not know the time.

\hypertarget{the-medium-the-machine-as-seen-by-its-members}{%
\subsection{The Medium: The Machine as Seen by Its
Members}\label{the-medium-the-machine-as-seen-by-its-members}}

This section specifies the semantics of the medium---the shape the
machine presents to its members. Section 3.3 specifies the runtime that
drives it.

\hypertarget{three-membranes-actor-to-channel-to-space}{%
\subsubsection{\texorpdfstring{Three Membranes: Actor \(\to\) Channel
\(\to\)
Space}{Three Membranes: Actor \textbackslash to Channel \textbackslash to Space}}\label{three-membranes-actor-to-channel-to-space}}

The machine is nested in three membranes.

\textbf{Actor: behavior.} An actor has a \texttt{kind}, a \texttt{text},
and a set of bound handles. Actors are the only things in the machine
that act: an actor receives one message and produces zero or more
actions and a reply (§3.3). Its \texttt{text} is its entire endowment.
For a program actor, \texttt{text} is source code; for a door, it is an
address.

\textbf{Channel: boundary.} A channel contains an append-only ledger, a
table of registered actors, one receptionist, and zero or more doors. A
channel has no code. It is pure structure; only the \texttt{text} of its
actors supplies behavior. In a machine description, a channel is one
structural node.

\textbf{Space: individual.} A Space comprises channels, the topology of
their doors, and one runtime instance. It is the smallest unit that
\(\Omega\) can spawn. Everything owned by a channel---ledger, member
table, and configuration---is quiescent data. The loop that moves it is
shared by all channels at Space level and belongs to none of them;
\texttt{Ω.spawn(channel)} has no entry point. A machine is a Space
together with its runtime. Throughout this paper, ``one machine'' means
one Space.

Membranes do not skip levels. An actor does not know what lies beyond
its channel; a channel does not know what else lies in its Space; a
Space does not know what else inhabits \(\Omega\). Everything a level
knows about its exterior enters as a message through its boundary
(Figure 2).

\begin{figure}
\centering
\includegraphics[width=1\textwidth,height=\textheight]{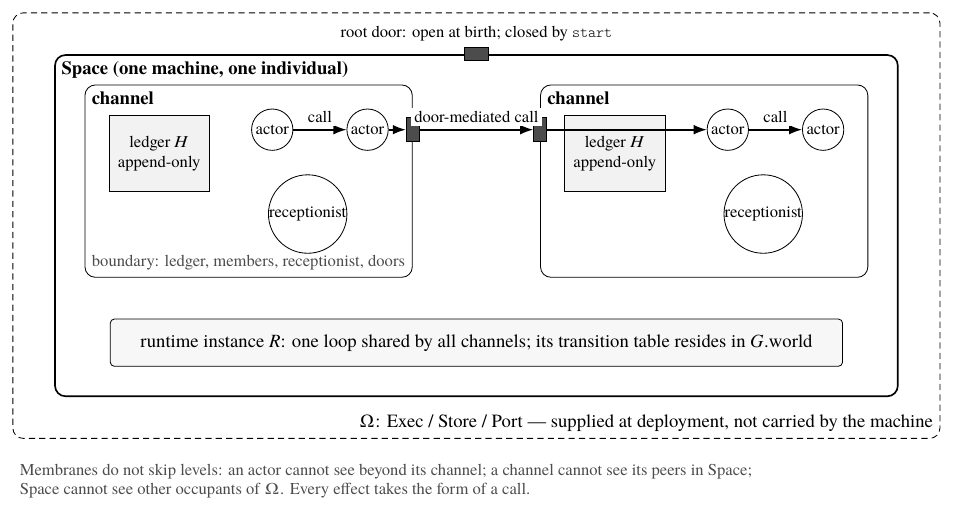}
\caption{Three membranes. Actors reside in channels, channels reside in
a Space, and \(\Omega\) lies outside the Space. Each channel contains an
append-only ledger, a member table, a receptionist, and doors. One
runtime instance \(R\) is shared by all channels at Space level. Every
effect is a call; calls across channels pass through doors. The root
door is open at birth and closes after \texttt{start}.}
\end{figure}

\hypertarget{call-is-the-only-action}{%
\subsubsection{Call Is the Only Action}\label{call-is-the-only-action}}

A member can do exactly one thing to the machine:

\[
\operatorname{call}(\text{address},\ \text{body}) \longrightarrow \text{return value}.
\]

There is no shared memory, no global variable, and no bypass around the
ledger. The address space is:

\begin{longtable}[]{@{}
  >{\raggedright\arraybackslash}p{(\columnwidth - 4\tabcolsep) * \real{0.3333}}
  >{\raggedright\arraybackslash}p{(\columnwidth - 4\tabcolsep) * \real{0.3333}}
  >{\raggedright\arraybackslash}p{(\columnwidth - 4\tabcolsep) * \real{0.3333}}@{}}
\toprule\noalign{}
\begin{minipage}[b]{\linewidth}\raggedright
Address
\end{minipage} & \begin{minipage}[b]{\linewidth}\raggedright
Meaning
\end{minipage} & \begin{minipage}[b]{\linewidth}\raggedright
Return value
\end{minipage} \\
\midrule\noalign{}
\endhead
\bottomrule\noalign{}
\endlastfoot
a tag such as \texttt{file} or \texttt{U} & a member of the current
channel & the member's reply \\
\texttt{0} & the read address of the medium & ledger rows or the member
table \\
the tag of a door & an outgoing path (§3.2.4) & empty \\
\texttt{syscall} plus a verb & a bound request that changes form or acts
on the world (§3.3.1) & receipt or rejection \\
\end{longtable}

A message addressed to nothing is discarded. A message addressed to a
member invokes that member synchronously, and the member's return value
becomes the return value of \texttt{call}. Nested calls form a call
stack, the machine's only control flow.

Reading address \texttt{0} does not change form, is open to every
member, and requires no binding. It recognizes two words.
\textbf{\texttt{show}} returns ledger rows: what have I experienced?
\textbf{\texttt{who}} returns the current member table: what am I now? A
third question---what ought I to be?---is answered by an organ (§3.4.2).
Together, the three questions locate in this substrate the problem left
unanswered by the 1948 model (§2.1): history, actual form, and
prescribed form each acquire a callable address.

Consider a member used repeatedly in §4. \texttt{file} is a program
actor whose body protocol is
\texttt{read\ \textless{}path\textgreater{}} or
\texttt{write\ \textless{}path\textgreater{}\textbackslash{}n\textless{}content\textgreater{}}.
When a member \texttt{X}, during one invocation, performs
\texttt{call("file",\ "read\ notes.txt")}, three rows appear in its
ledger:

\begin{Shaded}
\begin{Highlighting}[]
\NormalTok{msg  \{ seq k,   from X,    to file, body "read notes.txt", run j \}}
\NormalTok{step \{ seq k+1, actor file, ...,                            run j \}}
\NormalTok{msg  \{ seq k+2, from file, to X,    body "\textless{}file content\textgreater{}", run j \}}
\end{Highlighting}
\end{Shaded}

\texttt{X} receives the return value and continues. It does not know who
wrote \texttt{file} or when it was installed; the address is resolved at
the instant of the call. \texttt{file} is a part produced by the machine
itself (§4), and its caller cannot tell.

\hypertarget{the-ledger-h-to-be-entered-is-to-become-history}{%
\subsubsection{\texorpdfstring{The Ledger \(H\): To Be Entered Is to
Become
History}{The Ledger H: To Be Entered Is to Become History}}\label{the-ledger-h-to-be-entered-is-to-become-history}}

Each channel has one append-only ledger; the collection of all ledgers
in one machine is \textbf{\(H\)}. Rows fall into four classes.
\textbf{Morphology rows} (\texttt{place} and \texttt{retire}) record the
installation of a member at an address---including its complete
\texttt{text}---or the retirement of a member. Folding these rows yields
the channel's current form, implements \texttt{who}, and supplies the
basis for restart (§3.3.2). \textbf{Message rows} (\texttt{msg}) record
requests crossing a call boundary, nonempty replies, and receipts,
according to the address class (§3.3.1, transition 3). The \texttt{up}
message written by \(R\) to the receptionist of the first channel when a
machine wakes is also a \texttt{msg} row. \textbf{Step rows}
(\texttt{step}) close an invocation by recording the triggering sequence
number, every call it issued, and any error. \textbf{Boundary rows}
(\texttt{down}) record the completion of a legal shutdown and occur only
in the first channel's ledger (§3.3.1).

The first two laws of the medium are stated here; §3.3.4 establishes
their status.

\textbf{First, the medium stamps \texttt{from}.} No member can sign as
another. A message entering from outside is signed by the medium with
the corresponding door. In a ledger, who spoke is as trustworthy as what
was said.

\textbf{Second, entry is history.} A row appended to a ledger records an
event that has occurred. It is neither withdrawn nor rewritten. Every
recoverable ``current state'' is a fold over history, not a parallel
record. What has not entered a ledger has not happened to the machine.
Local variables, intermediate values, and remote exchanges inside one
actor invocation are absent from the ledger and do not belong to machine
history; only crossings of the call boundary do. The exogenous inputs of
§3.1.3 now have a location. Every nondeterministic value crossing the
boundary is preserved verbatim: history records not ``what was computed
then,'' but ``what was said then.''

The same boundary separates another class of effects. A member may touch
the host world during an invocation by reading a file or accessing a
network. Such actions are not actions on the machine: they are neither
recorded nor inherited. The machine's membrane is a membrane of
\textbf{organization and ledger}, not a computational sandbox (§1.3). An
effect becomes an effect of the machine only by becoming a ledger row.

\hypertarget{doors-the-only-kind-that-points-outward}{%
\subsubsection{Doors: The Only Kind That Points
Outward}\label{doors-the-only-kind-that-points-outward}}

A door is an actor whose \texttt{text} is not behavior but an
\textbf{address}. It is the only \texttt{kind} in the machine whose
\texttt{text} points outside its channel. A message sent to a door is
delivered unchanged to the endpoint named by its \texttt{text}, with
provenance stamped by the medium. A door emits no output and interprets
no content; it only transports. \textbf{The topology of a machine is the
set of \texttt{text} values of all its doors.} Two channels are
connected when their ledgers contain reciprocal doors. The far side may
be another channel in the same machine, a human, or another machine; the
door obeys the same rule. The plus sign in \(E=A+B+C+D\) (§2.1) now has
a definition: two organs are joined when their ledgers contain doors to
one another.

The distinction between a door and a member survives identical
implementations. A member is function: it receives a message and
produces actions within its channel; removing it removes a component. A
door is connection: it produces nothing and only transports; removing it
removes an edge. The same remote object installed as a member is
answering here; installed as a door, it is answering next door.

\hypertarget{the-way-in}{%
\subsubsection{The Way In}\label{the-way-in}}

The outside world has one way to act on a machine: \textbf{inbox \(\to\)
receptionist}. Every channel has an inbox, the receiving side of
\(\Omega\). The medium turns each arrival into a \texttt{msg} row, signs
it with the door corresponding to its origin, and addresses it to the
channel's \textbf{receptionist}---the member marked as such in the
morphology table. The statement ``every action on a channel is in its
ledger'' therefore holds for external actions as well.

There is one special door: the \textbf{root door}, which belongs to the
Space rather than to any channel. \textbf{The root door is open if and
only if no ledger contains a message row.} While open, it accepts two
kinds of input: syscalls that create form, and the first message. That
first message is both the machine's first heartbeat and the act that
closes the door. All subsequent input through the root door is ignored,
leaving the inbox as the machine's only interface to the world. Birth
(§3.4.4) determines who acted while the root door was open and what the
first message contained. The root door is open only before birth.

\hypertarget{the-runtime-r}{%
\subsection{\texorpdfstring{The Runtime
\(R\)}{The Runtime R}}\label{the-runtime-r}}

In the 1948 model, the mechanism that moves \(A\), \(B\), and \(C\)
belongs to the substrate physics, not to the machine or its description
(§2.1). On the new substrate, that physics is carried by a
\textbf{runtime \(R\)}: message delivery, member invocation, and ledger
append.

\(R\) is not \(\Omega\). The host knows nothing of channels, while the
entire purpose of \(R\) is to drive them; it arrives with the machine on
a blank host. Nor is \(R\) a member of the machine: it drives every
member, including the organs of construction (§3.4). It resides in a
root field of the machine description called \texttt{world} (§3.3.5). It
is inherited by every machine but contains no information about any one
machine. To the machine, \(R\) is physics: no operation in a member's
medium interface reads or modifies the current \(R\), and every member
is born immersed in it. To \(\Omega\), \(R\) is software. This is the
internal half of the environment contract (§2.2).

The definition of \(R\) is \textbf{a small state space and a transition
table invariant under renaming of content}. It has operational
semantics---what moves where---but no intentional semantics. It does not
know whether a step is construction or conversation.

\hypertarget{description-language-state-and-transitions}{%
\subsubsection{Description Language, State, and
Transitions}\label{description-language-state-and-transitions}}

A \textbf{legal \(G\)} belongs to the input domain of \(A\) (§3.4.1): it
is a description that \texttt{realize} can construct completely. Its
syntax is:

\begin{longtable}[]{@{}
  >{\raggedright\arraybackslash}p{(\columnwidth - 4\tabcolsep) * \real{0.3333}}
  >{\raggedright\arraybackslash}p{(\columnwidth - 4\tabcolsep) * \real{0.3333}}
  >{\raggedright\arraybackslash}p{(\columnwidth - 4\tabcolsep) * \real{0.3333}}@{}}
\toprule\noalign{}
\begin{minipage}[b]{\linewidth}\raggedright
Field
\end{minipage} & \begin{minipage}[b]{\linewidth}\raggedright
Contents
\end{minipage} & \begin{minipage}[b]{\linewidth}\raggedright
Constraint
\end{minipage} \\
\midrule\noalign{}
\endhead
\bottomrule\noalign{}
\endlastfoot
\texttt{world} & three texts: \(\omega\)-bind, loader, and \(R\) &
contains no information about this machine (§3.3.5) \\
\texttt{channels} & ordered list; each item contains a name, member
table, and receptionist & names are unique within the machine; the first
channel has a receptionist \\
member & \(\texttt{kind}\in\{\texttt{program},\texttt{door}\}\),
\texttt{text},
\(\texttt{bind}\subseteq\{\texttt{syscall},\texttt{spawn},\texttt{stop}\}\),
and \texttt{tag} & program \texttt{text} is source; door \texttt{text}
is an endpoint; instantiability and conformance are not syntactic
conditions (§§3.3.4, 3.4.3) \\
\texttt{peers} & pairs of channel names & each pair creates one door at
each endpoint \\
\end{longtable}

Legal \(G\) denotes the description language. The \(G_t\) present in a
machine state (§3.4.2) is one datum written in that language; whether it
remains in the language is itself a property of the state (§3.5).

The \textbf{recoverable state of \(R\)} consists of one append-only
ledger per channel, one cursor per actor indicating the last processed
event, and one offset per inbox indicating the last consumed line. All
three are reconstructed by folding the ledgers. Volatile member state
\(\Sigma\) is not included (§3.3.2). An \textbf{event} is a \texttt{msg}
row in a ledger, addressed to some address and not yet marked by a
\texttt{run}.

The complete transition relation of \(R\) is:

\begin{enumerate}
\def\labelenumi{\arabic{enumi}.}
\tightlist
\item
  When a line arrives in an inbox, append a \texttt{msg} for the
  channel's receptionist; \texttt{from} is the door pointing back to the
  sender's endpoint.
\item
  While the root door is open---that is, while no ledger contains a
  \texttt{msg} row---accept morphology-writing syscalls and the first
  message. Appending that first message closes the door (§3.2.5).
\item
  For an event, invoke the member at its address once. The invocation
  finishes before the next event is processed (run to completion). On
  completion, append a \texttt{step} row and advance the cursor; record
  an exception in the \texttt{step}. The machine has no other source of
  concurrency. Calls within the invocation are recorded by address
  class: a member address produces one request row and, for a nonempty
  reply, a second row; address \texttt{0}, a syscall, or a verb produces
  one fact or receipt row; a door returns empty; a nonexistent address
  produces no \texttt{msg} row and appears only in the call frame of the
  \texttt{step}.
\item
  A morphology syscall appends a morphology row; a world verb performs a
  host action, as listed below.
\item
  \texttt{stop} appends one \texttt{down} row after the current
  invocation completes, then exits \(R\).
\item
  When \(R\) wakes through \(\Omega.\operatorname{spawn}\) and the
  ledgers already contain a \texttt{msg}, fold every ledger and append a
  \texttt{msg\ up} to the receptionist of the first channel. Redeliver
  any event whose cursor was not advanced.
\item
  Killing the process produces no \texttt{down}. On the next wake, an
  unmatched \texttt{start} or \texttt{up} after the last \texttt{down}
  records an abnormal termination.
\end{enumerate}

An actor is a \textbf{resident function}. It is instantiated once when
its \texttt{place} row is folded, mounted at its address, and remains
there until retirement. Each subsequent message invokes \texttt{run(m)}
on the same object; the return value is the reply. \texttt{call} is
injected once at instantiation and is constant for that member. There
are two kinds:

\begin{longtable}[]{@{}
  >{\raggedright\arraybackslash}p{(\columnwidth - 4\tabcolsep) * \real{0.3333}}
  >{\raggedright\arraybackslash}p{(\columnwidth - 4\tabcolsep) * \real{0.3333}}
  >{\raggedright\arraybackslash}p{(\columnwidth - 4\tabcolsep) * \real{0.3333}}@{}}
\toprule\noalign{}
\begin{minipage}[b]{\linewidth}\raggedright
\texttt{kind}
\end{minipage} & \begin{minipage}[b]{\linewidth}\raggedright
Instantiation, once at placement
\end{minipage} & \begin{minipage}[b]{\linewidth}\raggedright
For each message
\end{minipage} \\
\midrule\noalign{}
\endhead
\bottomrule\noalign{}
\endlastfoot
program & \texttt{Exec.load(text,\ \{call,\ me,\ channel\})}: execute
source once to define \texttt{run} & \texttt{run(m)}; return value
becomes the reply \\
door & none (\texttt{text} is an endpoint address) & deliver unchanged
through \texttt{Port.send}, stamp provenance, and return empty \\
\end{longtable}

Requests that change morphology or act on the world are answered
directly by \(R\) and must be granted through the member's \texttt{bind}
field in \(G\). A request from an unbound member is discarded:

\begin{longtable}[]{@{}
  >{\raggedright\arraybackslash}p{(\columnwidth - 6\tabcolsep) * \real{0.2500}}
  >{\raggedright\arraybackslash}p{(\columnwidth - 6\tabcolsep) * \real{0.2500}}
  >{\raggedright\arraybackslash}p{(\columnwidth - 6\tabcolsep) * \real{0.2500}}
  >{\raggedright\arraybackslash}p{(\columnwidth - 6\tabcolsep) * \real{0.2500}}@{}}
\toprule\noalign{}
\begin{minipage}[b]{\linewidth}\raggedright
Group
\end{minipage} & \begin{minipage}[b]{\linewidth}\raggedright
Request
\end{minipage} & \begin{minipage}[b]{\linewidth}\raggedright
Action by \(R\)
\end{minipage} & \begin{minipage}[b]{\linewidth}\raggedright
Receipt
\end{minipage} \\
\midrule\noalign{}
\endhead
\bottomrule\noalign{}
\endlastfoot
morphology (\texttt{syscall}) & \texttt{channel.create\ name} & create
ledger and inbox & \texttt{name} \\
& \texttt{channel.add.actor}
\texttt{name\ kind\ text\ {[}bind{]}\ {[}tag{]}} & allocate a unique tag
and append a \texttt{place} row containing the \textbf{complete text} &
\texttt{channel/tag} \\
& \texttt{channel.retire.} \texttt{actor\ name/tag} & append a
\texttt{retire} row & \texttt{channel/tag} \\
world verbs & \texttt{spawn\ package} & \texttt{Ω.spawn(package)}: start
a new Space & process handle \\
& \texttt{stop} & append \texttt{down} after the current invocation,
then exit this Space & none \\
\end{longtable}

There is no third \texttt{kind}. \(L\) is also a program. Its
\texttt{text} is its entire loop: endpoint, prompt, ledger-view
assembly, framing of its output into calls, and continuation from
returned values. The medium sees only a sequence of calls and one reply.
\(R\) knows neither Python nor models; \(\Omega\) contains nothing
specific to \(L\).

\hypertarget{folding-and-restart-recovering-form-not-state}{%
\subsubsection{Folding and Restart: Recovering Form, Not
State}\label{folding-and-restart-recovering-form-not-state}}

The current form of a channel is not a second record. It is obtained by
folding its ledger: replay every \texttt{place} and \texttt{retire} row
to obtain the member table. This is the implementation of \texttt{who}
(§3.2.2). Restart therefore requires nothing beyond the ledgers:

\[
\text{restart} = \text{fold again} = \text{instantiate everything again}.
\]

On wake, \(R\) sends \texttt{up} to the receptionist of the first
channel. Events whose cursors were not advanced are redelivered; members
use their ledgers to suppress repeated effects.

A resident function may retain internal state \(\Sigma\) between
invocations---variables, caches, unfinished thoughts. \(\Sigma\) is not
in the ledger, which records crossings of the call boundary rather than
the inside of an invocation (§3.2.3), and it is reset by restart. A
living machine is \((G,H,\Sigma)\); \textbf{the individual is
\((G,H)\)}. It remains the same machine without preserving the same
volatile scene. The guarantee is \textbf{recovery of form, not recovery
of state}: folding \(H\) reconstructs actual morphology and
reinstantiates every member still present in it. It does not restore the
instant before failure. Whether the machine remains runnable,
constructible, or reproductive is a separate property (§3.5). A member
that must survive restart reconstructs itself by calling
\texttt{call("0",\ "show")}; that policy belongs to its \texttt{text},
not to \(R\).

\hypertarget{r-is-blind-to-function-and-organization}{%
\subsubsection{\texorpdfstring{\(R\) Is Blind to Function and
Organization}{R Is Blind to Function and Organization}}\label{r-is-blind-to-function-and-organization}}

\(R\) processes form, not meaning. It reads exactly three things:
addresses, \texttt{kind}, and \texttt{text} as a parameter to that kind.

\begin{longtable}[]{@{}
  >{\raggedright\arraybackslash}p{(\columnwidth - 2\tabcolsep) * \real{0.5000}}
  >{\raggedright\arraybackslash}p{(\columnwidth - 2\tabcolsep) * \real{0.5000}}@{}}
\toprule\noalign{}
\begin{minipage}[b]{\linewidth}\raggedright
Vocabulary known to \(R\)
\end{minipage} & \begin{minipage}[b]{\linewidth}\raggedright
Vocabulary unknown to \(R\)
\end{minipage} \\
\midrule\noalign{}
\endhead
\bottomrule\noalign{}
\endlastfoot
addresses, tags, doors & constructor, registry, author \\
\texttt{kind}: program or door & who may install a member, or whether an
installation counts \\
messages, ledgers, append, delivery, steps, receipts & \texttt{realize},
\texttt{pack}, \texttt{decl} \\
\texttt{text} as a parameter & the meaning of \texttt{text} \\
\end{longtable}

The condition is executable. Rename every organizational term in
\(G\)---rename \texttt{c0} to \texttt{x7}, for example---and rewrite
source equivalently. The behavior of \(R\) is unchanged, and the two
histories are isomorphic under the renaming. Violations are easy to
recognize: \texttt{R} contains \texttt{if\ name\ ==\ "c0"}; \(R\)
recognizes a ``construction request'' and performs it for a member;
\(R\) validates whether an object is a legal machine; or the scheduler
favors one member. Each places the proposition to be established---that
construction is performed by the machine---inside its physics. The
privilege of the construction and registry members (§3.4) is therefore
\textbf{organizational}. Their syscall bindings are written in \(G\),
and \(R\) binds them accordingly. Physically, they are equal to every
other member.

\hypertarget{the-laws-and-their-status}{%
\subsubsection{The Laws and Their
Status}\label{the-laws-and-their-status}}

The medium has three laws, the first two already stated in §3.2.3:

\begin{enumerate}
\def\labelenumi{\arabic{enumi}.}
\tightlist
\item
  \textbf{Append only, with \texttt{from} stamped by the medium.}
  History cannot be rewritten and provenance cannot be forged.
\item
  \textbf{Every action passes through a message.} No member can reach
  another around the ledger.
\item
  \textbf{Step fairness.} If \(R\) continues to run, the target has not
  retired, every invocation terminates, and no \texttt{stop} occurs,
  then every recorded message is eventually delivered. \(R\) neither
  favors, discards, nor reorders it.
\end{enumerate}

They are laws because they reside in \(R\), which is physics relative to
an individual. Every member---including the author---acts only through
\texttt{call}, and \texttt{call} is itself defined by these laws. Within
this medium, a violation is \textbf{inexpressible}. This is the finite
governance surface of §1.3: every organizational effect passes through
\texttt{call}, so authorization, audit, revocation, and termination each
have a definite location.

The boundary of the laws must be equally explicit. \textbf{Anything
interrupted midway is discarded.} If an invocation crashes halfway
through, everything not yet recorded---\(\Sigma\), local variables, an
unfinished row---does not belong to the machine. Effects already
recorded are history. The machine may therefore be left with a
half-modified morphology; this is damage, not contradiction, and repair
belongs to its organs (§3.4; §4.6). Undefinedness begins outside the
medium: when \(H\) is no longer a ledger that \(R\) can fold, or when
the host violates \(\Omega\), the physics required to describe the
machine no longer exists. \(R\) needs neither transactions nor two-phase
commit; it guarantees only that the ledger is truthful history.
\textbf{Redelivery is not replay.} Folding reconstructs morphology; an
event whose cursor was not advanced is delivered again. Whether a
previously recorded partial effect should be suppressed is decided by
the member from the ledger. Full replay of the machine's history is not
part of the model; §4.7 describes its cost.

\hypertarget{world-fixed-for-an-individual-variable-across-a-lineage}{%
\subsubsection{\texorpdfstring{\texttt{world}: Fixed for an Individual,
Variable Across a
Lineage}{world: Fixed for an Individual, Variable Across a Lineage}}\label{world-fixed-for-an-individual-variable-across-a-lineage}}

The description \(G\) has a root field named \textbf{\texttt{world}}
containing three texts: \(\omega\)-bind, which implements the \(\Omega\)
contract on a host; the loader, which defines directory layout and
startup; and \(R\), the transition table itself. Its status is captured
by two questions:

\begin{longtable}[]{@{}
  >{\raggedright\arraybackslash}p{(\columnwidth - 4\tabcolsep) * \real{0.3333}}
  >{\raggedright\arraybackslash}p{(\columnwidth - 4\tabcolsep) * \real{0.3333}}
  >{\raggedright\arraybackslash}p{(\columnwidth - 4\tabcolsep) * \real{0.3333}}@{}}
\toprule\noalign{}
\begin{minipage}[b]{\linewidth}\raggedright
\end{minipage} & \begin{minipage}[b]{\linewidth}\raggedright
Travels with the machine?
\end{minipage} & \begin{minipage}[b]{\linewidth}\raggedright
Contains information about this machine?
\end{minipage} \\
\midrule\noalign{}
\endhead
\bottomrule\noalign{}
\endlastfoot
\(\Omega\) & no---supplied at the destination & no \\
\texttt{world} & yes & no---identical for every machine in one world
version \\
the rest of \(G\) & yes & yes \\
\end{longtable}

The medium has no transition that changes the current \texttt{world}.
Members copy it without reading it; \(\Omega\) executes it; \(R\) does
not read \(G\). \texttt{world} is therefore fixed for an individual. It
remains variable across a lineage: an author may write
\texttt{world\textquotesingle{}} into the description of an offspring,
which is then born under new physics and accepted by a fixed-point test
(§4.4). Organs follow a different cadence. Members in \texttt{c0} and
\texttt{c1} may be added and retired within a generation, and a
completed replacement can be used immediately in the next reproduction.
One construction nevertheless uses already installed organs and one
fixed snapshot of \(G\) (§3.4.4).

The 1948 model's \(G\) does not contain transition rules; they belong to
its substrate (§2.1). Dalek's substrate is text. \texttt{world} travels
with every machine and evolves across a lineage, bringing physics itself
into heredity. This realizes the rule-heredity obligation of §1.2.

\hypertarget{organs-heredity-and-birth}{%
\subsection{Organs, Heredity, and
Birth}\label{organs-heredity-and-birth}}

Dalek distributes the four inherited roles of §2.1 across its organs and
adds a registry for heritable morphology (see note 3 in §2).
\textbf{\texttt{c0} carries \(A\), \(B\), and \(C\)}: \texttt{realize}
interprets \(G\) (\(A\)), \texttt{pack} copies \(G\) (\(B\)), and a
controller sequences the process (\(C\)). \textbf{\texttt{c1} is the
registry}, where \(G\) resides. In the 1948 model, the description is a
quiescent tape. Here \texttt{c1} folds its own ledger to produce the
description (§3.4.2). It is an additional organ---the place where growth
enters the description, as anticipated in §2.1. The registry is separate
from the constructor because prescribed and actual morphology must
remain independently comparable (§3.4.2). The granularity is consistent:
\(A\), \(B\), and \(C\) share \texttt{c0} because they jointly perform
one construction, while the object being compared occupies a ledger of
its own, so \(G_t\) is exactly the fold of one ledger. Quasi-quiescence
belongs to the snapshot returned by \texttt{decl}, which remains fixed
throughout \texttt{pack}, \texttt{spawn}, and \texttt{start} (§3.4.4).
\textbf{\texttt{c2} carries \(D\).} Organs are organization rather than
physics. \(R\) does not recognize \texttt{c0}, \texttt{c1}, or
\texttt{c2}; all their privilege consists of \texttt{bind} fields in
\(G\). \texttt{c0} is the constructor only because members implementing
\texttt{realize} reside there. Rename it or move those members to
another channel and the machine still runs.

The minimal machine \texttt{\{c0,c1\}} already closes reproduction. Its
description can be copied but not extended, because it has no author.
Adding \texttt{c2} yields the complete machine: it can maintain itself,
inherit, and vary into new capabilities. The following sections assemble
the three organs, then define birth, reproduction, and identity.

\hypertarget{c0-abc-the-only-interpreter-of-g}{%
\subsubsection{\texorpdfstring{\texttt{c0\ =\ A+B+C}: The Only
Interpreter of
\(G\)}{c0 = A+B+C: The Only Interpreter of G}}\label{c0-abc-the-only-interpreter-of-g}}

Construction has a dead half and a live half.
\textbf{\texttt{pack(G)\ →\ P}} is dead: it writes \texttt{G.world} into
files and places \texttt{G.json} beside them. It neither interprets
\(G\) nor generates anything, and the same description always produces
the same package. This is \(B\). \textbf{\texttt{realize}} is live: the
running \texttt{c0} reads the structure of \(G\) and issues syscalls
item by item, either through the root door to form an offspring or
locally to grow the present machine. This is \(A\). \textbf{No other
component interprets \(G\)}; every other hand that touches it merely
copies.

The central operation of \texttt{realize} constructs one organ from one
channel description. Constructing a machine means constructing its
organs in order and then installing doors according to the topology. A
channel is never copied in isolation. To ``copy a channel'' is to
construct another organ from the same description; it remains
construction.

The separation of \(A\) and \(B\) is visible at field level. A
description contains two classes of fields. \texttt{realize} interprets
\textbf{structure}---which channels exist, each member's kind, the
receptionist, and door destinations---but only transports \textbf{text},
which says what each member does and is later interpreted by \(R\). When
\texttt{c0} moves a piece of source, it does not know its meaning. \(A\)
reads organization, not content.

Universality lies in the accepted domain of \texttt{realize}: it works
for \textbf{every legal \(G\)}, not merely its own. A mechanism that
only copies itself is not \(A\).

\(C\) is another ordinary member of \texttt{c0}. Bound to the world
verbs, it sequences
\texttt{decl\ →\ pack\ →\ spawn\ →\ build\ →\ start}. Section 3.4.4
gives the complete path.

\hypertarget{c1-the-registry-and-the-split-of-morphology}{%
\subsubsection{\texorpdfstring{\texttt{c1}: The Registry and the Split
of
Morphology}{c1: The Registry and the Split of Morphology}}\label{c1-the-registry-and-the-split-of-morphology}}

Reproduction requires a quasi-quiescent description (§2.1), yet the
machine is alive and its running form is continuously affected by
messages. The resolution is to split \textbf{actual morphology} from
\textbf{heritable morphology}. Actual morphology \(M_t\) is the current
living structure obtained by \(R\) from all ledgers. Heritable
morphology \(G_t\) is the fold of \texttt{c1}'s ledger: the structure
the machine declares should be inherited and regenerated.

\texttt{c1} is a channel containing a registrar. It folds only its own
ledger, which receives through a door from \texttt{c0} three
declarations: \texttt{born} for items installed through the root door,
\texttt{placed}, and \texttt{retired} for actions performed by
\texttt{c0}. Thus

\[
\operatorname{decl}=G_0\oplus\texttt{placed}\ominus\texttt{retired}.
\]

\textbf{\texttt{c1} stores the commitments of \texttt{c0}, not the facts
of \(R\).} \(R\) records ``what I am now''; \texttt{c1} records ``what I
ought to be.''

They are related by a requirement: at rest,

\[
\pi(M_t) \cong G_t.
\]

The projection \(\pi\) is defined by provenance. It retains everything
installed by the constructor together with \texttt{c0} members installed
through the root door, and removes birth certificates, temporary doors,
and retired members. This relation is not an invariant of \(R\) but a
property maintained by \textbf{\texttt{c0}}. An organ holding
\texttt{syscall} can construct truth and register a lie; a corrupted
declaration is a mutation to be eliminated by descendant viability. The
membrane separates machine from world, not machine from its own organs.
That exposure is the substance of self-modification. During a morphology
change or after damage, \(M_t\) and \(G_t\) may temporarily diverge.

\textbf{Self-maintenance} has the following operational definition.
Relative to a target \(Q\) fixed before repair begins, the machine uses
its own construction path to eliminate a deviation or restore \(Q\). For
morphology, \(Q=G_t\) and repair brings \(\pi(M_t)\) back to \(G_t\);
for a function, \(Q\) is a member specification; for a capability, \(Q\)
may be runnability or reproductive capacity. \(Q\) is supplied by \(G\),
prior history, or a specification entering through the membrane.
Autonomous generation and internal storage of the target are outside
this paper.\footnote{A \textbf{declared target} has two layers. A
  specification states the capability the machine ought to
  possess---construction, copying, networking, or reproduction---and its
  acceptance conditions. A description gives the concrete implementation
  text of each member. \(G\) carries only the latter. In this paper,
  specifications remain implicit in the four-role architecture of §2.1,
  task messages, and experimenter judgments. A specification therefore
  enters the machine through a door. In §4.5, recovery of \(C\) follows
  an incoming protocol specification rather than \(G\); the rewritten
  \(C\) has a different implementation but satisfies the same
  specification. The fourth liveness condition in §4.7 refers to the
  same fact. This external input is a substitute for a specification
  that has no internal location. Bringing a specification into the
  machine requires no new medium mechanism: a member holding a
  specification and a door carrying one are both, at medium level, a
  member and a message. The difference is whether the specification is
  inherited in \(G\). Making the specification a first-class object
  distinct from \(G\)---so that specification defines completeness and
  \(G\) is one implementation satisfying it---is left to future work.}
Loss of completeness, loss of reproductive capacity, and a \(G_t\)
outside the description language are all states reachable by legal
transitions. The machine still has \(G_t\), \(H\), and a foldable
morphology in such states. A return path need not exist. If repair
depends on an organ that is itself damaged, the machine stops in a
well-defined but irrecoverable state and retains its identity and
history (§4.7).

This yields the cycle

\[
G_t \longrightarrow M_t \longrightarrow H \longrightarrow G_{t+1}.
\]

\texttt{realize} unfolds a description into live morphology. Activity
enters \(H\). The subset registered through \texttt{c0} enters the
ledger of \texttt{c1}, which folds it into the next description.
Unregistered history remains in \(H\) and does not enter \(G\).
Description and morphology define one another around this cycle, every
edge of which can be inspected in the ledgers (Figure 3).

\begin{figure}
\centering
\includegraphics[width=1\textwidth,height=\textheight]{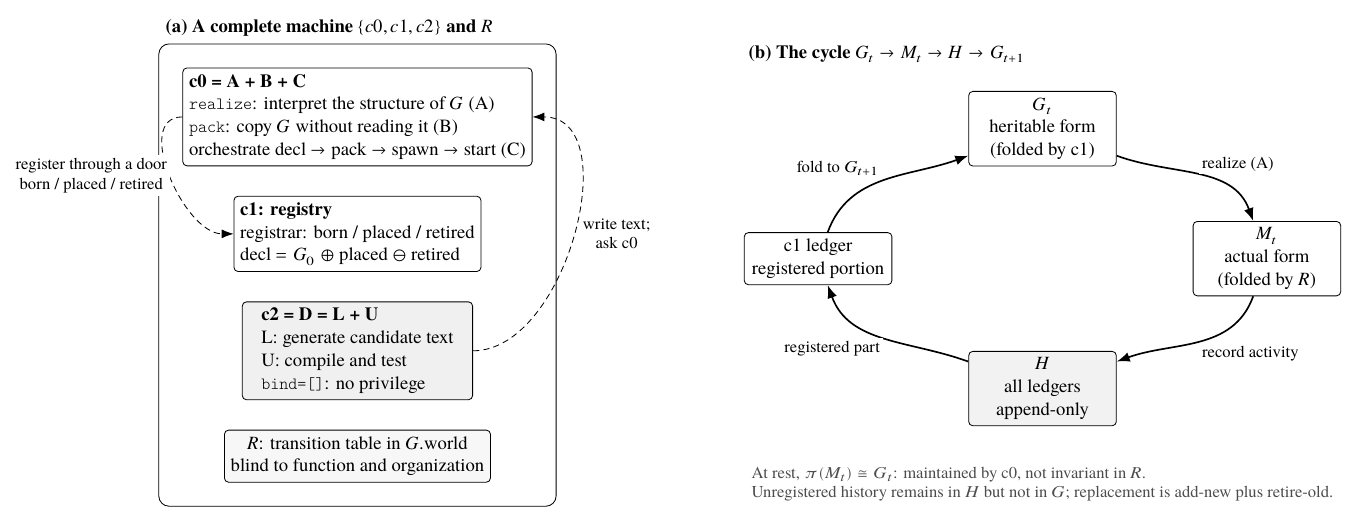}
\caption{(a) A complete machine: \texttt{c0} carries \(A\), \(B\), and
\(C\); \texttt{c1} is the registry; \texttt{c2\ =\ D\ =\ L+U} has no
bindings; and the transition table of \(R\) resides in \texttt{G.world}.
To change morphology, \texttt{c2} follows the same route as an external
party: write text and ask \texttt{c0}. (b) The cycle
\(G_t \to M_t \to H \to G_{t+1}\). \texttt{realize} unfolds the
description into actual morphology; activity is written to \(H\); the
portion registered through \texttt{c0} enters the ledger of \texttt{c1};
and \texttt{c1} folds it into the next description. At rest, \texttt{c0}
maintains \(\pi(M_t)\cong G_t\).}
\end{figure}

The syntax of change follows. There is \textbf{no \texttt{replace}}. A
\texttt{place} row cannot be rewritten because entry is history.
Modifying a member therefore means adding a new one and retiring the old
one. The new member receives a new physical address and may inherit the
old logical tag. Retirement appends a row and erases no history: the
member remains forever in history and disappears only from current
morphology.

\hypertarget{c2-d-lu-the-author-is-not-a-privileged-part}{%
\subsubsection{\texorpdfstring{\texttt{c2\ =\ D\ =\ L+U}: The Author Is
Not a Privileged
Part}{c2 = D = L+U: The Author Is Not a Privileged Part}}\label{c2-d-lu-the-author-is-not-a-privileged-part}}

The minimal machine cannot create a new member because it has no author.
\texttt{c2} supplies one. \textbf{\(L\)} generates: its \texttt{text} is
the entire agent loop, including ledger-view assembly, the remote model
call, and framing model output as individual calls. \textbf{\(U\)}
compiles: it turns candidate text into a part that can be installed. In
the first implementation, it receives candidate source plus tests,
instantiates the candidate in-process using the \textbf{same Exec} as
\(R\), runs the tests, and returns a result. Universality belongs to
\(L\) and Exec: what can be computed is determined by Exec, and what can
be written by \(L\); \(D\) adds neither. What \(D\) manufactures is
membership. It takes something \(L\) and Exec can already do and
constitutes it as a machine member: named and addressable, present in
\(H\) and \(G\), callable by any member, reinstantiated after restart,
copied verbatim during reproduction, and changed between versions by
adding one member and retiring another. Capability growth is the
accumulation of members, not an increase in underlying computational
power.

Neither component closes the loop alone. \textbf{\(L\) alone} produces
descriptions but supplies no internal criterion for distinguishing a
runnable part from one that merely resembles one; reliability would
remain an assumption about the author rather than a property of the
machine. \textbf{\(U\) alone} is an interpreter, which the machine
already possesses, and creates nothing new. Together they close the loop
proposal \(\to\) test \(\to\) revision \(\to\) installation. The inner
loop of \textbf{variation and selection enters the membrane}: the
machine decides whether a candidate runs; whether it is worth retaining
remains external. \textbf{Self-evolution} has the following operational
definition: a selected change enters \(G\) and is inherited along the
lineage. It is orthogonal to self-maintenance. If a repair of \(Q\)
enters \(G\) as a new implementation, the same run instantiates both
properties (§4.1). The source of novelty remains outside the
machine---the remote endpoint of \(L\) or a human---while the
\textbf{organization} that produces descriptions lies within it. The
machine does not generate randomness; it organizes randomness, just as
it organizes time (§3.1.3).

\(U\) is \textbf{necessary but insufficient}. A candidate is tested
inside the \texttt{c2} process with the author's \texttt{call} and
capabilities. Once installed in its target channel, it meets different
neighbors and bindings. The inner loop rejects candidates that cannot
run at all; it cannot reject every candidate that fails only after
relocation, as §4.3 demonstrates. External selection remains
unavoidable. \(U\) and \(R\) share one Exec: the compiler inside the
machine is an instance of the machine's physics (§6).

\textbf{The author is not privileged.} Both \(L\) and \(U\) have empty
bindings (\texttt{bind\ =\ {[}{]}}, inspectable in \(G\)). To alter
machine morphology, \texttt{c2} follows the same path as an external
requester: write text and ask \texttt{c0}. \(D\) can therefore be
replaced; two instances may coexist; an offspring may carry a different
\(D\). \(L\) can write a new \(L\), which \texttt{c0} installs and \(G\)
transmits. \textbf{The agent loop itself is heritable, mutable
morphology.} Only \texttt{run(m)} and \texttt{call} are fixed by the
medium. The payload position contains an author, and the seat itself is
part of \(G\).

\hypertarget{birth-closure-becomes-an-event}{%
\subsubsection{Birth: Closure Becomes an
Event}\label{birth-closure-becomes-an-event}}

How does a machine begin? There is no boot image preinstalled from
\(G\). \(\Omega.\operatorname{spawn}\) starts \(R\) with the root door
open and every ledger empty; the machine waits.

The lineage begins with a human. A human constructs the first machine
through the root door using the same syscalls, temporarily performing
\(A\), \(B\), and \(C\); sending \texttt{start} and closing the door is
the act of \(C\). Every subsequent generation is produced by organs
inside the machine. Birth has six steps, with each side recording one
half:

\begin{enumerate}
\def\labelenumi{\arabic{enumi}.}
\tightlist
\item
  The parent obtains \texttt{G\ =\ c1.decl()}.
\item
  \(C\) packages it: \(P'=\operatorname{pack}(G)\) (\(B\) copies without
  reading).
\item
  \(C\) starts a process: \(\Omega.\operatorname{spawn}(P')\). The
  offspring's \(R\) starts with an open root door and waits.
\item
  Through the root door, the parent's \texttt{realize} constructs
  \textbf{only \texttt{c0}}: it creates the channel, installs every
  \texttt{c0} member by moving \texttt{text} verbatim, and finally
  installs a door back to the parent. This birth certificate is absent
  from \(G\). The corresponding rows in the offspring are signed
  \texttt{by=\_root}.
\item
  Through the root door, \(C\) sends the first message:
  \texttt{start\textbackslash{}n\textless{}G\textgreater{}}. \textbf{The
  door closes and the offspring detaches.} The parent's obligation ends.
\item
  Receiving \texttt{start}, the offspring's \texttt{realize} uses local
  syscalls to grow every remaining channel, member, and connection, with
  rows signed by itself. Development completes; the machine becomes
  quiescent and waits for an inbox.
\end{enumerate}

\begin{figure}
\centering
\includegraphics[width=1\textwidth,height=\textheight]{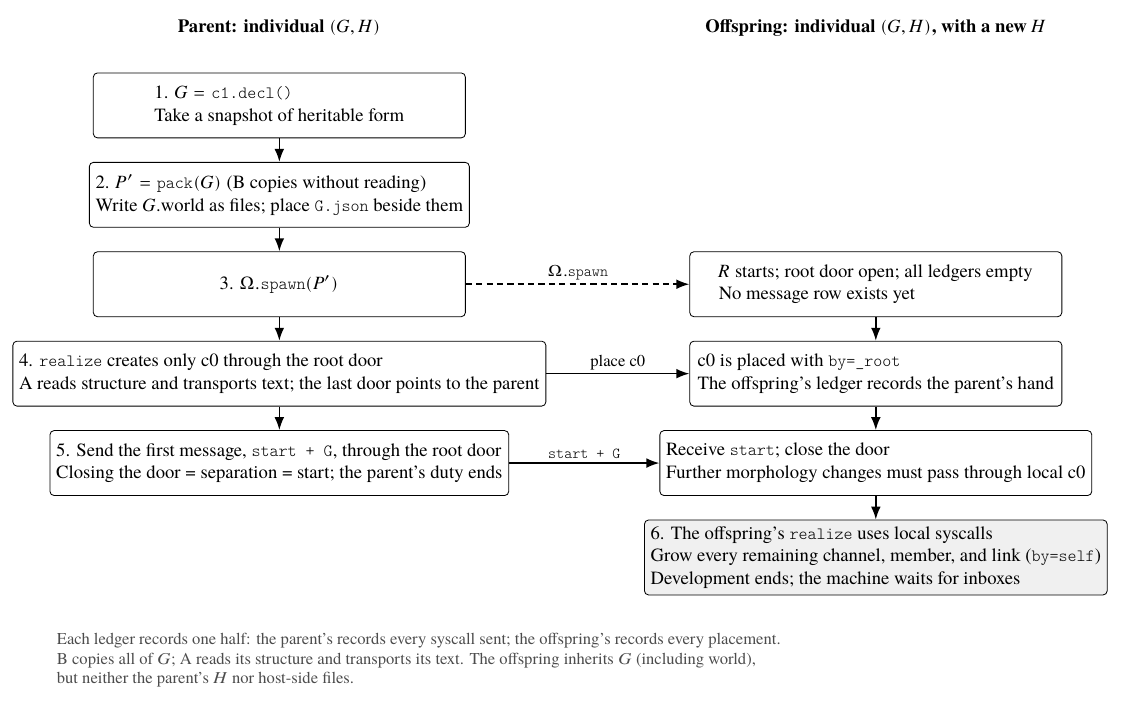}
\caption{Six steps of birth, half recorded in each ledger. Parent:
\texttt{decl}, \texttt{pack}, \(\Omega.\operatorname{spawn}\), construct
only \texttt{c0} through the root door, then send \texttt{start\ +\ G}
and close the door. Offspring: \(R\) starts and waits; \texttt{c0}
arrives with \texttt{by=\_root}; after \texttt{start} closes the door,
its own \texttt{realize} grows everything else with \texttt{by=self}.
The offspring inherits \(G\) and begins with an empty \(H\).}
\end{figure}

Three statements hold simultaneously in this path. \textbf{First}, the
parent never enters the offspring process; it only writes through the
offspring's root door. The offspring ledger exposes two hands: the root
and itself. The only organ not constructed by \texttt{c0} is \texttt{c0}
itself. \textbf{Second}, detachment, door closure, and \texttt{start}
are one event. While the door is open, external action determines
morphology and the machine is an object under construction. Once it
closes, morphology can change only through the machine's own
\texttt{c0}; the machine becomes a subject. \textbf{Closure changes from
an invariant to be guarded into a single event: birth.} \textbf{Third},
two quiescence conditions hold separately. Before \texttt{start}, the
offspring has no message row (§3.2.5), so nothing moves. The snapshot of
\(G\) obtained by \texttt{pack} remains fixed during copying. This
realizes quasi-quiescence (§2.1).

\hypertarget{reproduction-the-two-uses-of-g-become-field-operations}{%
\subsubsection{\texorpdfstring{Reproduction: The Two Uses of \(G\)
Become Field
Operations}{Reproduction: The Two Uses of G Become Field Operations}}\label{reproduction-the-two-uses-of-g-become-field-operations}}

The three reproductive operations \texttt{decl\ →\ pack\ →\ spawn} all
appear in the birth path. The \textbf{two uses of \(G\)} (§2.1) become
two treatments of two field classes. \texttt{pack} copies all of \(G\)
(\(B\) reads nothing), while \texttt{realize} reads structure and
transports text (\(A\) interprets organization but not content). \(B\)
gives the constructor no back door: the source of \texttt{c0} and
\texttt{c1} is copied from their two \texttt{text} fields in the
\textbf{description}, not from the code currently executing. The two
uses of the description apply to the constructor itself, and an observer
can compare package and description to check that nothing was omitted.

The output of \texttt{pack} is dead---source, not a process. \(A\)
constructs a quiescent machine, which \(C\) then starts (§2.1). Only
dead material can cross hosts: bytes travel over a network, processes do
not. Only dead material can be compared: the fixed-point test compares
two descriptions (§4.4). Reproduction is replayable from both sides. The
parent's ledger contains every syscall sent; the offspring's ledger
contains every syscall received.

\hypertarget{identity-individual-gh}{%
\subsubsection{\texorpdfstring{Identity: Individual
\(=(G,H)\)}{Identity: Individual =(G,H)}}\label{identity-individual-gh}}

The first addition beyond the 1948 model (§2.1) now has a definition:

\[
\text{individual}=(G,H)
\]

\(G\) answers ``what am I?'' and \(H\) answers ``which one am I?'' \(G\)
is copyable because copying is the purpose of a quasi-quiescent
description. \(H\) is not inherited: an offspring starts with an empty
history. Two machines with byte-identical \(G\) are indistinguishable in
morphology and diverge from the first row of their histories. Section
3.3.2 supplies the dynamical half of the definition: restart clears
\(\Sigma\) while preserving \((G,H)\), so the machine remains the same
individual.

The third question of §3.2.2 is also discharged. Introspection now has
three addresses. \texttt{show} returns the history of the current
channel---what have I experienced? \texttt{who} returns its actual
morphology---what am I now? \texttt{decl} returns the heritable
morphology \(G_t\) of the whole machine---what ought I to be? The whole
machine's actual morphology \(M_t\) is \(R\)'s fold over every ledger,
not the answer of one address. On this substrate, the question
unanswered by the 1948 model becomes three calls. No new axiom is
required beyond the existence of the ledger.

Names follow from identity rather than registering it. An external
naming authority is unnecessary: an offspring is the reproductive event
in its parent's ledger. Its package path lies beneath
\texttt{spawn/\textless{}name\textgreater{}} in the parent package, and
a grandchild nests beneath it. This is a lineage-local name, unique
within a lineage; offspring of two founders may share a name. Identity
remains \((G,H)\), and names are derived from ledgers. A host-global
counter instead causes collisions as soon as two machines merge,
allowing identity to leak in from the host rather than reside in
description and history.

\hypertarget{from-the-1948-model-to-dalek}{%
\subsubsection{From the 1948 Model to
Dalek}\label{from-the-1948-model-to-dalek}}

The following table accounts for what Dalek inherits from the
constructional model, what the new substrate rederives, and what the
agent-machine obligations add.

\begin{longtable}[]{@{}
  >{\raggedright\arraybackslash}p{(\columnwidth - 6\tabcolsep) * \real{0.2500}}
  >{\raggedright\arraybackslash}p{(\columnwidth - 6\tabcolsep) * \real{0.2500}}
  >{\raggedright\arraybackslash}p{(\columnwidth - 6\tabcolsep) * \real{0.2500}}
  >{\raggedright\arraybackslash}p{(\columnwidth - 6\tabcolsep) * \real{0.2500}}@{}}
\toprule\noalign{}
\begin{minipage}[b]{\linewidth}\raggedright
Relation
\end{minipage} & \begin{minipage}[b]{\linewidth}\raggedright
1948 constructional model (§2.1)
\end{minipage} & \begin{minipage}[b]{\linewidth}\raggedright
Dalek
\end{minipage} & \begin{minipage}[b]{\linewidth}\raggedright
Consequence
\end{minipage} \\
\midrule\noalign{}
\endhead
\bottomrule\noalign{}
\endlastfoot
retained & four-way division into \(A/B/C\)/payload; two uses of a
description; startup and detachment by \(C\); construction-time
quasi-quiescence & retained item by item & the hereditary constructional
core is inherited from §2.1 \\
replaced & \(A\) constructs cell by cell with a construction arm &
\texttt{realize} issues syscalls from \(G\) & the blueprint is already a
part; the arm collapses into a fold rule \\
replaced & universality means every automaton in the cellular medium &
universality means every legal \(G\) & universality is relative to the
legal description space of the medium \\
replaced & substrate is a 29-state cellular automaton & environment
contract is \(\Omega+R\): \(\Omega\) explicit, \(R\) inherited & the
substrate is replaceable and identity independent of host (§3.1.2) \\
replaced & the parent's \(A\) constructs the complete offspring & the
parent constructs only \texttt{c0}, copies \(G\), and sends
\texttt{start}; the offspring grows the rest & the parent's obligation
is minimized and the offspring's constructor is tested at birth \\
added & the description is a quiescent tape & \texttt{c1}: a registry
whose description is folded from a ledger & growth enters the
description (§3.4.2) \\
added & no history & \(H\): append-only, stamped ledgers & identity,
introspection, and recovery (§§3.2--3.3) \\
added & payload is inactive & \(D=L+U\), the author & mutation becomes
an organ rather than external noise \\
added & transition rules are absent from the description &
\texttt{world} is inherited inside \(G\) & physics enters the lineage
(§3.3.5) \\
\end{longtable}

The hereditary constructional core comes from the 1948 model (§2.1).
Dalek's machine architecture results from combining that core with the
definitional obligations (§1.2) and the properties of an agent-specific
substrate; the registry and \(D\) then close the cycles of growth and
variation on the completed medium. Dalek is therefore a new agent
machine containing the 1948 construction as one core, not a new
implementation of von Neumann's machine. Section 4 supplies a running
witness for each row.

\hypertarget{the-machine-against-the-criteria}{%
\subsection{The Machine Against the
Criteria}\label{the-machine-against-the-criteria}}

The four obligations of §1.2 now have the following concrete forms.

\textbf{Host boundary.} The account has four locations. Mechanisms
reside in \(G\) and \(\Omega\): \(G\) contains \texttt{world} and the
\texttt{text}, bindings, and receptionist marks of every member; \(R\)
resides in \texttt{G.world}; the privilege of an organ is encoded in the
\texttt{bind} fields of its members. The \(\Omega\) table contains three
capabilities and none of this paper's vocabulary. Facts reside in \(H\):
ledger rows witness every claimed property. Volatile execution state
resides in \(\Sigma\): it affects current behavior---a member's counter
may return to one after restart---but claims about identity, morphology
recovery, and cross-generational heredity do not require it to persist
(§3.3.2). Every mechanism affecting the properties at issue is either
text in \(G\) or a capability in \(\Omega\); there is no third location.

\textbf{Construction language.} Two kinds and three morphology syscalls
generate every actual form. \(R\) accepts no other kind, so none appears
in \(M_t\). A legal \(G\) in the description language (§3.3.1) uses the
same primitives. New capability production is unbounded because the
output of \(L\) has no inventory. Entry into the machine has two
moments. A \texttt{place} row enters actual morphology \(M_t\); a
\texttt{placed} declaration registered in \texttt{c1} enters heritable
morphology \(G_t\). Failure may occur between them, splitting \(M_t\)
from \(G_t\) (§3.4.2). In both places, the persistent form is text.
Generation is open-ended; installation has a finite language.

\textbf{Admissible transitions.} Legal state is delimited by the medium
laws (§3.3.4): ledgers append only; \texttt{from} is stamped by the
medium; rows have five kinds; morphology-row kinds are either program or
door; and the current receptionist has not retired. \(R\) can fold any
set of ledgers satisfying these conditions. The initial machine is the
first legal state produced by genesis (§3.4.4), and the seven rules of
§3.3.1 are its admissible transitions. Preservation follows from the
position of
\$R\texttt{:\ \$R\$\ is\ the\ sole\ ledger\ writer,\ and\ every\ row\ written\ by\ a\ transition\ satisfies\ the\ laws.\ A\ member\ can\ change\ constitution\ only\ through}call\texttt{.\ A\ morphology\ request\ from\ an\ unbound\ member\ never\ becomes\ an\ action\ of\ \$R};
an invalid request from a bound member becomes a rejection receipt, not
a morphology row. No action by a member can therefore produce an illegal
state. By induction, every state reachable from the initial machine
through admissible transitions is legal. The class is defined by medium
law, not completeness. Completeness, runnability, constructibility,
reproductive capacity, and membership of \(G_t\) in the description
language are properties of legal states and can be lost through legal
transitions (§3.4.2). Maintaining and restoring them belongs to the
organs (§4).

\textbf{Rule heredity.} The rules that produce a legal successor have
three layers, all contained in \(G\). \texttt{world} contains the
transition rules of the medium; the \texttt{text} of \texttt{c0} and
\texttt{c1} contains the rules of construction, copying, control, and
registration; the \texttt{text} of \(D\) contains the policy for
generating variations and selecting candidates. The first two define
structural legality, while \(D\) determines which changes to search.
Every layer is representable as text, constructible because
\texttt{realize} transports it, and heritable because \texttt{pack}
copies it verbatim. Replacement follows two cadences. \texttt{world} is
fixed within an individual and takes effect in a successor; organ
members can be replaced within a generation (§3.3.5). A particular
construction uses already installed organs and one fixed snapshot of
\(G\), so event order eliminates the circle of constructing oneself at
the same instant (§2.1).

\hypertarget{the-machine-in-operation-from-one-task-to-a-third-generation}{%
\section{The Machine in Operation: From One Task to a Third
Generation}\label{the-machine-in-operation-from-one-task-to-a-third-generation}}

Section 3.5 separates two layers of machine property. \(R\) preserves
legality of the medium; the organs maintain completeness, runnability,
constructibility, and reproductive capacity. This section supplies
organ-level witnesses in five classes: constitution and heredity of a
capability (§4.2), organization and repair of a population (§4.3),
replacement of rules across generations (§4.4), loss and recovery of a
member (§4.5), and crash, shutdown, and damage (§4.6). Each trace
changes a different property. The four \emph{selves} overlap on one
machine, and each has identifiable ledger rows; one run may instantiate
several at once (§4.1). The occupant of payload \(D\) is an external
general-purpose model, identified for each experiment. Tasks, diagnoses,
ticks, and spawn requests enter through doors. A human never bypasses a
door to rewrite machine organization. Fault injection---killing a
process or deleting a ledger---occurs on the host side as an event in
\(\Omega\), not as a machine transition (§3.3.1). A statement that ``the
machine did'' something refers to ledger rows, not to an author's
report.\footnote{Appendix B maps every ledger reference in this section
  to the evidence snapshot shipped with the paper. The original E1 and
  E2 ledgers predate tag addressing, so external receipts use numeric
  addresses such as \texttt{c2/5}; the current protocol uses a
  channel-local unique tag such as \texttt{c2/file} (§3.3.1). Original
  records are quoted unchanged.}

\hypertarget{four-forms-of-self}{%
\subsection{Four Forms of Self}\label{four-forms-of-self}}

Each form has an operational definition and an acceptance statement.

\begin{longtable}[]{@{}
  >{\raggedright\arraybackslash}p{(\columnwidth - 6\tabcolsep) * \real{0.2500}}
  >{\raggedright\arraybackslash}p{(\columnwidth - 6\tabcolsep) * \real{0.2500}}
  >{\raggedright\arraybackslash}p{(\columnwidth - 6\tabcolsep) * \real{0.2500}}
  >{\raggedright\arraybackslash}p{(\columnwidth - 6\tabcolsep) * \real{0.2500}}@{}}
\toprule\noalign{}
\begin{minipage}[b]{\linewidth}\raggedright
Property
\end{minipage} & \begin{minipage}[b]{\linewidth}\raggedright
Operational definition
\end{minipage} & \begin{minipage}[b]{\linewidth}\raggedright
Acceptance statement
\end{minipage} & \begin{minipage}[b]{\linewidth}\raggedright
Witness
\end{minipage} \\
\midrule\noalign{}
\endhead
\bottomrule\noalign{}
\endlastfoot
\textbf{self-maintenance} & Relative to a target \(Q\) fixed before
repair and supplied by \(G\), prior history, or an external
specification (§3.4.2), the machine uses its own construction path to
remove a deviation or restore \(Q\). Morphology:
\(\pi(M_t)\not\cong G_t\to\pi(M_t)\cong G_t\). Function: a member again
satisfies its specification. Capability: properties such as reproduction
are restored. Neither \(R\) nor \(\Omega\) changes. & A channel whose
ledger was deleted is reconstructed from \texttt{decl} (morphology, E3);
a nonconforming member is replaced through the machine's own syscall
path (function, E2-c); an erroneously retired reproductive member is
rewritten from its specification and reinstalled (capability, E5). &
§§4.3, 4.5, 4.6 \\
\textbf{self-evolution} & A selected variation enters \(G\) and is
inherited along the lineage (§3.4.3); the inner loop of variation and
selection lies within the membrane. & Faced with a missing part, the
author writes, compiles, and installs it; it enters \(G\) and is
inherited verbatim (E1, live model). A retired member is reimplemented
from its specification, installed, and inherited by two generations (E5,
live model). A change to \texttt{world} is rule heredity rather than
this property (E4, §4.4). & §§4.2, 4.5 \\
\textbf{self-reproduction} & \texttt{decl\ →\ pack\ →\ spawn} closes;
\texttt{decl(offspring)\ =\ decl(parent)} and the offspring can
reproduce again. \(G\), including \texttt{world}, is inherited;
host-side files and effects are not. & The offspring inherits the new
part verbatim but not the file it operated on. Three generations carry
the same rewritten member, and the parent and child each use it to
produce the next generation. & §§4.2, 4.5 \\
\textbf{self-organization} & Through inherited protocols and messages,
multiple machines grow and maintain their own local topology. No
external party writes the topology, and there is no shared morphology or
ledger. The coordinator is an ordinary replaceable member. & Three
machines each grow doors to the other two. An early message is lost and
the next heartbeat repairs the relation. After one machine is killed, a
neighbor wakes it from its own ledger. & §4.3 \\
\end{longtable}

The rows are not exclusive: E5 is simultaneously self-maintenance,
self-evolution, and self-reproduction. Figure 5 places the five
experiments on ledgers and generations.

\begin{figure}
\centering
\includegraphics[width=1\textwidth,height=\textheight]{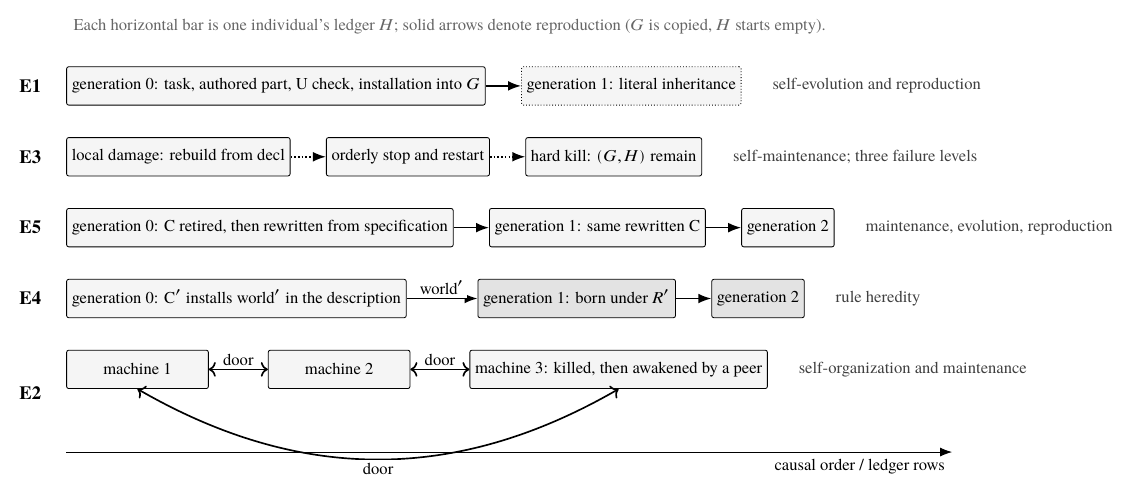}
\caption{Five experiments. Each horizontal bar is the ledger \(H\) of
one individual; ledger rows advance to the right, and solid arrows
denote reproduction. E1 traces one task and inheritance by an offspring.
E3 exercises three failure levels in one machine. E5 carries a rewritten
\(C\) across two generations. In E4, an offspring carrying
\texttt{world\textquotesingle{}} is born under new physics and
reproduces. In E2, three machines grow doors from an inherited protocol
and a neighbor wakes one after it is killed.}
\end{figure}

All four are mechanism claims. In every experiment, what to change, what
to repair, and whether the result is wanted are supplied from outside
the membrane as tasks, specifications, or diagnoses. The machine
provides the path by which change and repair occur and by which their
results are inherited. Autonomous choice of what to change remains
external (§5.3).

\hypertarget{one-task-end-to-end-e1}{%
\subsection{One Task, End to End (E1)}\label{one-task-end-to-end-e1}}

The apparatus is the smallest complete machine that can write:
\texttt{\{c0,c1,c2\}}, with \texttt{c2\ =\ \{L,U,two\ doors\}}. The
remote endpoint of \(L\) is \texttt{deepseek-chat}. The machine is born
and becomes quiescent. The experimenter sends one input through a door:

\begin{Shaded}
\begin{Highlighting}[]
\NormalTok{task}
\NormalTok{write hello into notes.txt, then read it back}
\end{Highlighting}
\end{Shaded}

The machine has no member named \texttt{file}. Its ledger records:

\begin{longtable}[]{@{}
  >{\raggedright\arraybackslash}p{(\columnwidth - 2\tabcolsep) * \real{0.5000}}
  >{\raggedright\arraybackslash}p{(\columnwidth - 2\tabcolsep) * \real{0.5000}}@{}}
\toprule\noalign{}
\begin{minipage}[b]{\linewidth}\raggedright
\texttt{seq}
\end{minipage} & \begin{minipage}[b]{\linewidth}\raggedright
Ledger event
\end{minipage} \\
\midrule\noalign{}
\endhead
\bottomrule\noalign{}
\endlastfoot
5 & \texttt{task} enters from a door to \(L\); the first invocation
begins. \\
6--7 & \(L\) calls \texttt{call("0","show")} and
\texttt{call("0","who")}, reading history and morphology---the
introspection of §3.4.6, recorded as fact rows. \\
8→10 & \(L\) sends candidate source to \(U\): \texttt{U\ test} v1 →
\texttt{result\ 1} because the first line is parsed incorrectly. \\
11→13 & v2 → \texttt{result\ 1} because the test delimiter is missing
and the tests do not run. \\
14→16 & v3 → \texttt{result\ 0}. \\
17 & Through a door, \(L\) asks \texttt{c0}:
\texttt{add\ c2\ program\ tag=file\ iface=…}; fourteen lines of source
enter the row. \\
19 & The first invocation closes with step frame
\texttt{{[}0,\ 0,\ U,\ U,\ U,\ c0{]}} and no return value. \\
20--21 & \(R\) installs and instantiates \texttt{c2/5}; the
\texttt{placed} receipt returns through the door and begins a second
invocation. \\
24--29 &
\texttt{L\ call("file","write\ notes.txt\textbackslash{}nhello")} →
\texttt{written}; then \texttt{call("file","read\ notes.txt")} →
\texttt{hello}. The invoked object is the newly installed machine part,
not \(U\). \\
30 & \texttt{done} returns through the door to the requester. \\
\end{longtable}

The path takes approximately ninety seconds and six HTTP rounds across
two invocations. Every mechanism participates once. The task enters
through a door (§3.2.5). The author assembles introspection (§3.2.2).
Two failed candidates and one successful candidate pass through \(U\)
(§3.4.3), leaving two rounds of debugging in the ledger. A new
capability has one way into the machine: door \(\to\) \texttt{c0}
\(\to\) syscall \(\to\) installation by \(R\) \(\to\) registration
\(\to\) receipt as the next invocation (§3.4.1). The machine then
invokes the part it has just grown to complete the task.

The experiment had a failed predecessor, E1-a. Its prompt used the
wording of a mechanism test, and the model followed it literally: it
wrote a one-off part with the task hard-coded in the function body,
installed but never invoked it, and reported \texttt{done} directly. The
correction touched neither \(R\), \(\Omega\), nor the construction
organs \texttt{c0} and \texttt{c1}. Only a passage in the \texttt{text}
of \(L\) changed: when a capability is missing, add a general part, use
it after \texttt{placed}, and resolve paths relative to the current
working directory. Repeating the run produced the table above. The two
runs differed only in text inside \(G\) and behaved entirely
differently: the behavior of the author is text in \(G\) (§3.4.3).

\(U\) could already write the file because it can execute arbitrary
Python; E1-a wrote it inside \(U\), and the machine gained nothing. E1-b
performs the same operation but produces a member. \texttt{file} has a
name and address; its \texttt{place} row carries all source into \(H\);
\texttt{placed} registers it in \(G\); any member may call it; restart
reinstantiates it; and reproduction copies it verbatim. E1 does not
increase the computational expressiveness of \(L\) or Exec. It
\textbf{constitutes} a capability as part of the machine: a latent
operation of a universal executor becomes named, addressable,
persistent, reusable, and heritable (§3.4.3). The difference is
irrelevant to a one-off script and decisive for a system of many parts
that must be maintained, copied, and moved over time (§1.1).

After \texttt{spawn}, \texttt{c2/5} in the offspring is the same
\texttt{file}, byte for byte, and
\texttt{decl(offspring)\ =\ decl(parent)}. The offspring directory
contains no \texttt{notes.txt}. Calling \texttt{read\ notes.txt} on its
\texttt{file} returns \texttt{FileNotFoundError}; after \texttt{write},
a file appears in the offspring and the parent's remains unchanged. The
part is in \(G\) and the file is on the host. Description, including
\texttt{world}, is inherited; host-side effects are not. This is the
hereditary counterpart of the membrane in §3.2.3.

E1 satisfies the acceptance statements for self-evolution---the selected
change enters \(G\) and is inherited verbatim---and the first half of
self-reproduction---verbatim inheritance without inheritance of host
effects. Its limits are explicit. The prompt names \texttt{file},
guiding what to build but not how to build or install it. The experiment
is a single run. E1-a and E1-b compare ``text as policy''; they are not
controlled repetitions.

\hypertarget{a-population-e2-self-organization-and-self-maintenance}{%
\subsection{A Population (E2): Self-Organization and
Self-Maintenance}\label{a-population-e2-self-organization-and-self-maintenance}}

The apparatus is the same machine with \(L\) replaced by
\texttt{deepseek-v4-pro}. The task specifies the behavior of two
members: a \texttt{hub}, which acknowledges registrations and broadcasts
its member table, and a \texttt{reporter}, which pings peers on each
heartbeat and invokes \texttt{spawn} when one disappears. Four protocol
words are supplied; implementation is left to the author. Time arrives
from outside the membrane as specified in §3.1.3, with the experimenter
writing \texttt{tick} into an inbox.

Installation, from specification to running members, takes eight minutes
in the author ledger of \texttt{d0}:

\begin{longtable}[]{@{}
  >{\raggedright\arraybackslash}p{(\columnwidth - 2\tabcolsep) * \real{0.5000}}
  >{\raggedright\arraybackslash}p{(\columnwidth - 2\tabcolsep) * \real{0.5000}}@{}}
\toprule\noalign{}
\begin{minipage}[b]{\linewidth}\raggedright
\texttt{seq}
\end{minipage} & \begin{minipage}[b]{\linewidth}\raggedright
Ledger event
\end{minipage} \\
\midrule\noalign{}
\endhead
\bottomrule\noalign{}
\endlastfoot
9 & A 1,160-character task specifying the two members enters from door
\texttt{me} to \(L\). \\
10--11 & \(L\) reads \texttt{show} and \texttt{who}. \\
12→14 & \(L\) asks \(U\) to run
\texttt{os.path.abspath(\textquotesingle{}.\textquotesingle{})} →
\texttt{result\ 0:\ /tmp/t1-real}, obtaining the allowed working
directory. \\
15→20 & Source for each member passes once through \(U\)
(\texttt{result\ 0}, \texttt{result\ 0}); both external iterations
finish without \texttt{err}. \\
21 & Through a door to \texttt{c0}:
\texttt{add\ c3\ program\ in\ tag=hub}; the complete hub source enters
the row. \\
23 & \texttt{add\ c4\ program\ in\ bind=spawn\ tag=reporter};
approximately 3,000 characters of reporter source enter the row. \\
25, 27 & \texttt{peer\ c0\ c3}; \texttt{peer\ c0\ c4}, creating the
connections. \\
29 & \texttt{add\ c4\ door\ tag=hub\ →\ file:/tmp/t1-real\#c3}, the
reporter's door to the hub. \\
31 & \texttt{done} returns through door \texttt{me}. \\
33 & The invocation closes with step frame
\texttt{{[}0,\ 0,\ U,\ U,\ U,\ c0,\ c0,\ c0,\ c0,\ c0,\ me{]}}. \\
34--50 & \texttt{placed} receipts return in turn (\texttt{c3/1},
\texttt{c4/1}, \texttt{c0/6}, \texttt{c3/2}, \texttt{c0/7},
\texttt{c4/2}, \texttt{c4/3}), each becoming a new invocation. \\
\end{longtable}

The experimenter then sends \texttt{tick} to an inbox, completing the
hub loop. Through doors, the experimenter asks \texttt{c0} to produce
\texttt{d1} and \texttt{d2}. Both offspring inherit the hub and
reporter. The number of births is externally chosen; topology thereafter
is grown by the machines themselves. The reporter ledger of \texttt{d1}
gives the clearest view. Every member in it is inherited, and the
original task was never sent to it:

\begin{longtable}[]{@{}
  >{\raggedright\arraybackslash}p{(\columnwidth - 2\tabcolsep) * \real{0.5000}}
  >{\raggedright\arraybackslash}p{(\columnwidth - 2\tabcolsep) * \real{0.5000}}@{}}
\toprule\noalign{}
\begin{minipage}[b]{\linewidth}\raggedright
\texttt{seq}
\end{minipage} & \begin{minipage}[b]{\linewidth}\raggedright
Ledger event
\end{minipage} \\
\midrule\noalign{}
\endhead
\bottomrule\noalign{}
\endlastfoot
1--3 & Morphology present at birth and inherited from \(G\):
\texttt{reporter} (\texttt{in}, \texttt{bind=spawn}), a door to
\texttt{c0}, and a door to \texttt{hub}. \\
4 & \texttt{start}, beginning the first invocation. \\
7 & Through its hub door, the reporter registers:
\texttt{hello\ file:…/spawn/d1\#c4}. \\
10 & The hub broadcasts
\texttt{peers\ \textless{}three\ endpoints\textgreater{}}. \\
13, 15 & For each unfamiliar endpoint, the reporter asks \texttt{c0}
through a door to \texttt{add\ c4\ door}. \\
19, 22 & Two new doors enter the ledger. Topology has grown (§3.2.4:
topology is the set of doors). \\
25, 28 & The reporter sends \texttt{ping} through each new door. \\
31--32 & \texttt{pong} returns. All three machines communicate six
seconds after the birth of \texttt{d1}. \\
11, 18 & Interleaved earlier: a \texttt{ping} from \texttt{d2} arrives
before the return door exists and is signed \texttt{door}; the
reporter's \texttt{pong} at 20 and 23 is addressed to \texttt{door} and
discarded by the medium (§3.2.2). The next heartbeat repairs the
relation; recovery from a lost message is part of the protocol. \\
\end{longtable}

The three machines share neither ledger nor morphology. They are an
ecology, not one individual (§5). Coordination passes through the hub:
it collects registrations and broadcasts the member table, from which
each reporter grows local doors. The hub is an ordinary member installed
by the same path as \texttt{file} (§4.2), and it can be replaced or
retired. No external party writes topology; each machine grows its own
from an inherited protocol.

The experiment then introduces damage. After the experimenter sends
\texttt{SIGKILL} to \texttt{d0}, repeated ticks on \texttt{d1} fail to
invoke \texttt{spawn}. The defect is visible directly in the ledger:

\begin{longtable}[]{@{}
  >{\raggedright\arraybackslash}p{(\columnwidth - 2\tabcolsep) * \real{0.5000}}
  >{\raggedright\arraybackslash}p{(\columnwidth - 2\tabcolsep) * \real{0.5000}}@{}}
\toprule\noalign{}
\begin{minipage}[b]{\linewidth}\raggedright
\texttt{seq} (\texttt{d1/c4})
\end{minipage} & \begin{minipage}[b]{\linewidth}\raggedright
Ledger event
\end{minipage} \\
\midrule\noalign{}
\endhead
\bottomrule\noalign{}
\endlastfoot
94 & Before repair, the \texttt{tick} step frame contains bare
\textbf{\texttt{spawn}}. The author had written
\texttt{call("spawn",\ d)} with two arguments, while this medium
requires the verb and argument on the same first line (Appendix A). The
parser therefore sees only \texttt{spawn}, cannot identify a member, and
silently discards the request. Failure detection succeeds; rescue is
never sent; no receipt appears. \\
\end{longtable}

One spelling error disables remote maintenance for the whole machine:
the ABI is a real surface of variation. Repair follows the machine's own
path (E2-c). Diagnosis occurs outside and enters the \texttt{c2} of
\texttt{d1} through a door, together with the old source and the fault
location. Approximately seventy seconds later, repair is recorded across
two ledgers:

\begin{longtable}[]{@{}
  >{\raggedright\arraybackslash}p{(\columnwidth - 2\tabcolsep) * \real{0.5000}}
  >{\raggedright\arraybackslash}p{(\columnwidth - 2\tabcolsep) * \real{0.5000}}@{}}
\toprule\noalign{}
\begin{minipage}[b]{\linewidth}\raggedright
\texttt{seq}
\end{minipage} & \begin{minipage}[b]{\linewidth}\raggedright
Ledger event
\end{minipage} \\
\midrule\noalign{}
\endhead
\bottomrule\noalign{}
\endlastfoot
\texttt{c2:25} & The repair task enters, carrying the old source and
diagnosis. \\
\texttt{c2:28} & \(L\) writes a new reporter and requests through a
door: \texttt{add\ c4\ program\ in\ bind=spawn\ tag=reporter}. Its
source visibly contains \textbf{\texttt{call("spawn\ "\ +\ d)}}. \\
\texttt{c2:31} & Receipt: \texttt{placed\ c4/6}. \\
\texttt{c2:34→37} & \texttt{retire\ c4/1\ →\ retired}. The rule ``no
replace; add new, retire old'' (§3.4.2) is applied to a live member. \\
\texttt{c4:111–112} & The new reporter is installed with \texttt{in},
taking over reception; the old one retires. \\
\texttt{c4:133} & \texttt{tick}, the first heartbeat of the new
reporter. \\
\texttt{c4:138} & Spawn receipt: \texttt{/tmp/t1-real\ pid=2705088}.
\texttt{d0} wakes by folding its own ledger (§3.3.2); its neighbor has
pressed the switch. \\
\texttt{c4:146} & The step frame now contains
\textbf{\texttt{spawn\ /tmp/t1-real}}, a one-word difference from
sequence 94. \\
\texttt{c4:147} & \texttt{pong} returns and the loop is restored. \\
\end{longtable}

E2 satisfies the acceptance statements for self-organization: machines
grow mutual doors, heartbeats repair lost contact, and a neighbor wakes
a killed machine from its own ledger, so population-level organization
participates in maintenance. It also satisfies the statement for
functional self-maintenance: the machine replaces a nonconforming member
through its own syscall path. The task asks for an implementation of a
given specification; invention is not tested. Diagnosis and evaluation
occur outside and enter through a door. This is a single run. One driver
race was initially confused with a failure of the subject and separated
by the ledgers.\footnote{The original ledgers of E2 were generated
  before lifecycle semantics were tightened. Restart events were then
  injected into every channel; they are now recorded only in the first
  channel (§3.3.1). The original evidence is not rewritten. The
  conclusions in this section do not depend on which ledger contains the
  physical lifecycle event; current semantics are transitions 5--7 in
  §3.3.1 and the traces of §4.6.}

\hypertarget{the-world-e4-an-offspring-with-a-new-r-reproduces}{%
\subsection{\texorpdfstring{The World (E4): An Offspring with a New
\(R\)
Reproduces}{The World (E4): An Offspring with a New R Reproduces}}\label{the-world-e4-an-offspring-with-a-new-r-reproduces}}

E4 is a deterministic mechanism experiment using a stub author. It
witnesses the claim of §3.3.5 that \texttt{world} is fixed for an
individual and variable across a lineage. The variation enters through
the member responsible for packaging. \texttt{C\textquotesingle{}} is
the original \(C\) plus one idempotent patch: before \texttt{pack},
modify the ledger writer in \texttt{G.world} to stamp every row with the
constant field \texttt{"w":\ 2}. The following messages are literal.
\(R\), \(\Omega\), \texttt{syscall}, and \(A\) remain unchanged.

\begin{longtable}[]{@{}
  >{\raggedright\arraybackslash}p{(\columnwidth - 4\tabcolsep) * \real{0.3333}}
  >{\raggedright\arraybackslash}p{(\columnwidth - 4\tabcolsep) * \real{0.3333}}
  >{\raggedright\arraybackslash}p{(\columnwidth - 4\tabcolsep) * \real{0.3333}}@{}}
\toprule\noalign{}
\begin{minipage}[b]{\linewidth}\raggedright
Step
\end{minipage} & \begin{minipage}[b]{\linewidth}\raggedright
Event in the medium
\end{minipage} & \begin{minipage}[b]{\linewidth}\raggedright
Assertion over ledger and disk
\end{minipage} \\
\midrule\noalign{}
\endhead
\bottomrule\noalign{}
\endlastfoot
1 & Through a door: \texttt{retire\ c0/C}. \(C\) is not a receptionist
and may retire. & A \texttt{retire} row appears. \\
2 & Through a door: \texttt{add\ c0\ program\ tag=C\ bind=syscall,spawn}
plus source for \texttt{C\textquotesingle{}}. The same role takes over
the same logical address. & The old member is retired; the new member
has \texttt{tag=C}. \\
3 & Through a door: \texttt{spawn\ w2}. Birth follows the ordinary six
steps of §3.4.4, now performed by \texttt{C\textquotesingle{}}. & The
receipt contains a process identifier. \\
4 & --- & No parent ledger row contains \texttt{w}; an individual is
unchanged. \\
5 & --- & The offspring's \texttt{runtime.py} contains \texttt{"w":\ 2},
so \texttt{world} contains \(R'\). Every row of the offspring's
\texttt{c0} and \texttt{c1} ledgers carries \texttt{"w":\ 2}, because
\(R'\) wrote them. \\
6 & --- & The offspring's \(C\) is byte-identical to
\texttt{C\textquotesingle{}} modulo address allocation;
\texttt{decl(offspring).world} contains the mark. The variation entered
the registry and is heritable. \\
7 & Send \texttt{spawn\ w2g} to the offspring, testing the fixed point:
reproduction still works under \(R′\). & A grandchild is born and
develops; the receipt contains a process identifier. \\
8 & --- & The grandchild's \texttt{runtime.py} contains the mark, every
ledger row carries \texttt{"w":\ 2}, and its \(C\) is still
byte-identical to \texttt{C\textquotesingle{}}. The patch is idempotent
and does not accumulate. \\
\end{longtable}

One ledger witnesses the two cadences of rule heredity described in
§3.5. The member \texttt{C\textquotesingle{}} is installed within the
parent's lifetime at step 2 and used immediately for reproduction at
step 3. \texttt{world\textquotesingle{}} is written by
\texttt{C\textquotesingle{}} into the offspring's description and takes
effect only in the offspring at steps 4--5.

The ledger writer is selected for three reasons. It maximizes
observability: any row identifies which \(R\) wrote it and whether the
parent was affected. It maximizes harmlessness: no transition rule
changes, and machines under \(R\) and \(R′\) can reproduce one another,
so the experiment tests reachability rather than the quality of a new
runtime. It also makes idempotence decidable at step 8.

E4 satisfies the fourth obligation, rule heredity. Variation of
\texttt{world} is reachable through the ordinary member-installation
path; the individual remains fixed; the change takes effect in the next
generation; and hereditary closure holds. The self-evolution loop does
not participate. The variation is specified and written by the
experimenter, then installed through a door without \(L\) or \(U\). The
mark changes no transition rule. The path for rule heredity is
established; an end-to-end run in which \(D\) authors
\texttt{world\textquotesingle{}} has not been performed.

\hypertarget{a-member-e5-a-lost-c-is-rewritten-and-used-by-two-generations}{%
\subsection{\texorpdfstring{A Member (E5): A Lost \(C\) Is Rewritten and
Used by Two
Generations}{A Member (E5): A Lost C Is Rewritten and Used by Two Generations}}\label{a-member-e5-a-lost-c-is-rewritten-and-used-by-two-generations}}

Section 3.5 assigns completeness to the properties maintained by organs.
Losing a member does not kill the machine; it leaves the machine
temporarily incomplete. E5 removes \(C\), the member in \texttt{c0} that
carries the reproductive role and the only one of the three original
roles that can be removed by a legal transition. \(A\) is the
receptionist of \texttt{c0}, and \(R\) refuses to retire a current
receptionist; \(B\) is \texttt{pack} inside the process controlled by
\(C\).

During this run, the author did not retrieve the old source. The task
omitted it, and the ledger of \texttt{c2} exposed only \texttt{show} and
\texttt{who} for that channel. The old \texttt{text} of \(C\) remained
in the ledger \(H\) of \texttt{c0}, but retrieving it would have
required the receptionist of \texttt{c0} to send it through a door, an
interface the current \(A\) does not provide. The restored \(C\) differs
structurally from the original, satisfies the same protocol, and
supports reproduction in two generations. Re-creation is therefore a
behavioral conclusion rather than a claim about hidden access.

The live-model run E5-b uses \texttt{deepseek-v4-pro}. The task supplies
only the two-response protocol that \(C\) must implement and no old
source. The complete run takes four minutes and leaves one segment in
each of three ledgers.

Loss and rewriting, in the \texttt{c0} and \texttt{c2} ledgers of the
parent:

\begin{longtable}[]{@{}
  >{\raggedright\arraybackslash}p{(\columnwidth - 2\tabcolsep) * \real{0.5000}}
  >{\raggedright\arraybackslash}p{(\columnwidth - 2\tabcolsep) * \real{0.5000}}@{}}
\toprule\noalign{}
\begin{minipage}[b]{\linewidth}\raggedright
\texttt{seq}
\end{minipage} & \begin{minipage}[b]{\linewidth}\raggedright
Ledger event
\end{minipage} \\
\midrule\noalign{}
\endhead
\bottomrule\noalign{}
\endlastfoot
\texttt{c0:47–48} & Through a door, the experimenter sends
\texttt{retire\ c0/C}. The member carrying reproduction retires, and a
\texttt{retire} row is appended. \\
\texttt{c0:54–55} & \texttt{spawn\ kid}: \(A\) forwards the request to
\(C\), visible as step frame \texttt{{[}C{]}}, but \(C\) has retired.
There is no receipt and no directory. Reproductive capacity is gone. \\
\texttt{c2:9} & A 1,451-character repair task containing only the
protocol specification enters. \\
\texttt{c2:10–11} & \(L\) reads \texttt{show} and \texttt{who}. \\
\texttt{c2:12→14} & A 2,084-character candidate goes to \(U\):
\texttt{run\ →\ result\ 0}. \\
\texttt{c2:15} & Through a door:
\texttt{add\ c0\ program\ bind=syscall,spawn,stop\ tag=C}. All 2,112
characters of the new \(C\) enter the row. \\
\texttt{c2:17} & The invocation closes with step frame
\texttt{{[}0,\ 0,\ U,\ c0{]}}. \\
\texttt{c0:58} & \texttt{place\ tag=C,\ textlen=2112,\ by=1}. The
machine's own hand reinstalls the member. \\
\texttt{c2:18–21} & Receipt \texttt{placed\ c0/C} arrives as a new
invocation; \texttt{done} returns through a door. \\
\end{longtable}

Restored reproduction, still in the parent's \texttt{c0} ledger:

\begin{longtable}[]{@{}
  >{\raggedright\arraybackslash}p{(\columnwidth - 2\tabcolsep) * \real{0.5000}}
  >{\raggedright\arraybackslash}p{(\columnwidth - 2\tabcolsep) * \real{0.5000}}@{}}
\toprule\noalign{}
\begin{minipage}[b]{\linewidth}\raggedright
\texttt{seq}
\end{minipage} & \begin{minipage}[b]{\linewidth}\raggedright
Ledger event
\end{minipage} \\
\midrule\noalign{}
\endhead
\bottomrule\noalign{}
\endlastfoot
66--67 & \texttt{spawn\ kid} is sent again. \(A\) forwards it to \(C\),
which now responds. \\
69→77 & \(C\) requests \texttt{decl}; \texttt{c1} folds its ledger and
returns a 48,147-character \(G\). \\
79 & Spawn receipt: \texttt{spawn/kid\ pid=2873033}. The offspring
process starts with an open root door. \\
85→99 & \(C\) asks \(A\) to \texttt{build}. Six syscalls enter the
offspring's root door: create \texttt{c0}, add \(A\), add doors to
\texttt{c1} and \texttt{c2}, \textbf{add \(C\) including its own
2,218-character text}, and add the birth-certificate door (§3.4.4, step
4). \\
100 & Through the root door, \(C\) sends \texttt{msg\ c0\ start\ +\ G},
74,815 characters. This first message closes the door and detaches the
offspring. \\
102 & \texttt{spawned\ /tmp/restore-real/spawn/kid\ door=kid}. \\
\end{longtable}

Heredity, in the \texttt{c0} ledgers of offspring and grandchild:

\begin{longtable}[]{@{}
  >{\raggedright\arraybackslash}p{(\columnwidth - 2\tabcolsep) * \real{0.5000}}
  >{\raggedright\arraybackslash}p{(\columnwidth - 2\tabcolsep) * \real{0.5000}}@{}}
\toprule\noalign{}
\begin{minipage}[b]{\linewidth}\raggedright
\texttt{seq}
\end{minipage} & \begin{minipage}[b]{\linewidth}\raggedright
Ledger event
\end{minipage} \\
\midrule\noalign{}
\endhead
\bottomrule\noalign{}
\endlastfoot
\texttt{kid:1–5} & Five rows installed through the root door with
\texttt{by=\_root}: \(A\), doors to \texttt{c1} and \texttt{c2},
\textbf{\(C\) with \texttt{textlen=2112}, inherited verbatim}, and the
birth-certificate door. \\
\texttt{kid:6→35} & \texttt{start} arrives with \(G\). The offspring's
own hand (\texttt{by=1}) grows \texttt{c1}, \texttt{c2}, and every
connection, completing development (§3.4.4, step 6). \\
\texttt{kid:38,43} & \texttt{spawn\ grand} enters; \(A\) forwards it to
the inherited \(C\). \\
\texttt{kid:70,93} & Receipts \texttt{spawn/grand\ pid=2877373} and
\texttt{spawned\ …/kid/spawn/grand}; the lineage is written in the path
(§3.4.6). \\
\texttt{grand:4} & \texttt{place\ tag=C,\ textlen=2112,\ by=\_root}: the
third generation carries the same 2,112-character member. \\
\texttt{grand:6→35} & The same developmental sequence repeats. \\
\end{longtable}

Three generations carry the \(C\) written by the author. Parent and
offspring each use it to produce the next generation, and the grandchild
inherits it and completes development. This is the same acceptance shape
as the fixed point in §4.4: a new component is not restored until it
supports the next reproduction itself.

Three observations follow. \textbf{First, an unprivileged author creates
a privileged member.} \(L\) and \(U\) both have \texttt{bind\ =\ {[}{]}}
(§3.4.3). Their add request asks for \texttt{bind=syscall,spawn,stop},
and \texttt{c0} grants it. Privilege is conferred by \texttt{c0};
whether such a request should be approved is policy in \(A\), not
physics in \(R\). \textbf{Second, the restored \(C\) is different from
but equivalent to the original.} It locates the receptionist through
\texttt{who}, whereas the original used a heuristic; computes unhandled
requests as a set difference over receipts rather than by counting; and
handles nested paths omitted by the original. The protocol is preserved
and the implementation is new, supporting re-creation rather than backup
restoration. \textbf{Third, the structural rule ``do not retire the
current receptionist'' preserves receptionist positions in all three
channels.} \(A\), the registrar, and \(L\) are receptionists and cannot
be retired directly. \(C\) and \(U\) are the core members that can be
lost directly. \(R\) prevents the receptionist position from becoming
empty in one step but does not guarantee that a successor is usable.
Capability completeness remains an organ-level property (§3.5).

E5 satisfies the acceptance statements for the second half of
self-reproduction---three generations carry the member and two use it to
reproduce---self-evolution---the rewritten \(C\) enters \(G\) through
the inner loop and is inherited by two generations---and
capability-level self-maintenance---a retired reproductive member is
re-created from a specification and reproductive capacity returns. The
specification is supplied externally. Discovery of what is missing and
selection of what to restore remain outside. This is a single run.

\hypertarget{crash-shutdown-and-damage-e3}{%
\subsection{Crash, Shutdown, and Damage
(E3)}\label{crash-shutdown-and-damage-e3}}

Failure of one member is the lightest level. If an invocation raises an
exception, \(R\) records \texttt{err} in its \texttt{step}, returns
empty, and advances the cursor. An unreachable remote endpoint of \(L\)
likewise becomes one \texttt{err} row while the machine remains alive.
The two \texttt{result\ 1} values for candidate source in §4.2 belong to
a different category: \(U\) catches those failures inside its test and
returns them as normal replies. Its \texttt{step} contains no
\texttt{err}; failure exists only as a selection result in the inner
loop (§3.4.3).

A crash midway through an event is more severe. Section 3.3.4 defines
anything unrecorded as outside the machine and any recorded partial
effect as history. A syscall executes and appends its row immediately
within an invocation, while the enclosing \texttt{step} is appended only
when the invocation returns. A crash in between leaves a half-modified
morphology. This is a damaged state in the sense of §3.5. The ledger
remains truthful, and self-maintenance means bringing \(\pi(M_t)\) back
to \(G_t\). \(R\) needs neither transactions nor two-phase commit.

Replay cannot replace redelivery. \texttt{channel.add.actor} is not
idempotent: replaying one add creates a second member under a newly
allocated tag. Actual morphology then contains two while \(G\) records
one, and \(\pi(M_t)\cong G_t\) no longer holds. The extra member is
damage to be repaired by an organ. This is half the reason for
``redeliver, do not replay'' in §3.3.4; §4.7 supplies the other half.

The following traces come from E3, a deterministic mechanism experiment
with a stub author. Assertions are over ledgers and disk.

\textbf{Local damage and reconstruction.} The apparatus is the minimal
machine plus an echo member \texttt{x}; damage is introduced at the
host.

\begin{longtable}[]{@{}
  >{\raggedright\arraybackslash}p{(\columnwidth - 4\tabcolsep) * \real{0.3333}}
  >{\raggedright\arraybackslash}p{(\columnwidth - 4\tabcolsep) * \real{0.3333}}
  >{\raggedright\arraybackslash}p{(\columnwidth - 4\tabcolsep) * \real{0.3333}}@{}}
\toprule\noalign{}
\begin{minipage}[b]{\linewidth}\raggedright
Step
\end{minipage} & \begin{minipage}[b]{\linewidth}\raggedright
Event in the medium
\end{minipage} & \begin{minipage}[b]{\linewidth}\raggedright
Assertion
\end{minipage} \\
\midrule\noalign{}
\endhead
\bottomrule\noalign{}
\endlastfoot
1 & Through a door: \texttt{add\ c8\ program\ in\ tag=e};
\texttt{peer\ c0\ c8}. & The \texttt{c8} ledger contains a
\texttt{place} row; \texttt{c1} contains a \texttt{placed\ c8}
declaration; record \texttt{decl}. \\
2 & Stop the machine; delete the ledger file of \texttt{c8} on the host.
& --- \\
3 & Start the same package \(P\) with \(\Omega.\operatorname{spawn}\);
\(R\) folds. & \(M_t\) has no \texttt{c8}, because a channel exists iff
its ledger contains a \texttt{place} row. The first channel gains one
\texttt{up}. \\
4 & \(A\) receives \texttt{up}, sends \texttt{reconcile} to \texttt{c1},
and issues syscalls channel by channel against \texttt{decl}. &
Undamaged channels return \texttt{exists} and are skipped. \texttt{c8}
is reconstructed with the same text, tag, receptionist position, and
door to \texttt{c0}. \\
5 & --- & The ledger of \texttt{c8} begins empty: the channel is
reconstructed from its description, not replayed. \texttt{c1} does not
register a duplicate \texttt{placed\ c8}; its ledger has no \texttt{up}
or \texttt{down}, because physical lifecycle events appear only in the
first channel. \\
6 & --- & \texttt{decl} is unchanged from step 1; \(\pi(M_t)\cong G_t\)
is restored. \\
\end{longtable}

Repair is performed by the \texttt{text} of \(A\). Translating
\texttt{up} into \texttt{reconcile}, comparing the records, and
rebuilding are policy in \(A\); \(R\) only folds and emits \texttt{up}.
This witnesses morphological self-maintenance with \(Q=G_t\) (§3.4.2).

\textbf{Legal shutdown and wake.} The same apparatus runs as an
independent process.

\begin{longtable}[]{@{}
  >{\raggedright\arraybackslash}p{(\columnwidth - 4\tabcolsep) * \real{0.3333}}
  >{\raggedright\arraybackslash}p{(\columnwidth - 4\tabcolsep) * \real{0.3333}}
  >{\raggedright\arraybackslash}p{(\columnwidth - 4\tabcolsep) * \real{0.3333}}@{}}
\toprule\noalign{}
\begin{minipage}[b]{\linewidth}\raggedright
Step
\end{minipage} & \begin{minipage}[b]{\linewidth}\raggedright
Event in the medium
\end{minipage} & \begin{minipage}[b]{\linewidth}\raggedright
Assertion
\end{minipage} \\
\midrule\noalign{}
\endhead
\bottomrule\noalign{}
\endlastfoot
1 & Birth and development; send \texttt{hi} to \texttt{x}. & The ledger
of \texttt{x} contains \texttt{hi} and its \texttt{step}. \\
2 & Through a door to \texttt{c0}: \texttt{stop}; \(A\to C\to\) the stop
verb. & The \texttt{c0} ledger records \texttt{stopping}; the first
channel records \texttt{down} as its final row; the process exits; the
ledger of \texttt{x} has no \texttt{down}. \\
3 & While stopped, send \texttt{late} to \texttt{x}. & It remains in the
inbox. \\
4 & Start the same \(P\) with \(\Omega.\operatorname{spawn}\). &
Lifecycle rows in the first channel are \texttt{down,\ up};
\texttt{late} enters the ledger and is processed. \\
5 & Send \texttt{stop} through a door again. & Lifecycle rows are
\texttt{down,\ up,\ down}; the \texttt{place} rows and \texttt{decl} of
\texttt{x} are unchanged; cursors persist across processes. \\
\end{longtable}

\textbf{Hard kill.} The apparatus is unchanged.

\begin{longtable}[]{@{}
  >{\raggedright\arraybackslash}p{(\columnwidth - 4\tabcolsep) * \real{0.3333}}
  >{\raggedright\arraybackslash}p{(\columnwidth - 4\tabcolsep) * \real{0.3333}}
  >{\raggedright\arraybackslash}p{(\columnwidth - 4\tabcolsep) * \real{0.3333}}@{}}
\toprule\noalign{}
\begin{minipage}[b]{\linewidth}\raggedright
Step
\end{minipage} & \begin{minipage}[b]{\linewidth}\raggedright
Event in the medium
\end{minipage} & \begin{minipage}[b]{\linewidth}\raggedright
Assertion
\end{minipage} \\
\midrule\noalign{}
\endhead
\bottomrule\noalign{}
\endlastfoot
1 & Send \texttt{late} to \texttt{x} and kill \(R\) after the message is
recorded but before it runs. & --- \\
2 & Start again. & There is \texttt{up} without a matching
\texttt{down}, identifying the previous abnormal termination. The cursor
is recovered by folding; \texttt{late} remains pending and runs again
under at-least-once delivery; \texttt{decl} is unchanged. \\
3 & With pending messages \texttt{a} and \texttt{b}, send \texttt{stop}
through a door. & They are abandoned immediately; \texttt{down} is the
final row. \\
4 & Start again. & \texttt{a} and \texttt{b} run, producing
\texttt{echo:a} and \texttt{echo:b}. \\
\end{longtable}

External \texttt{SIGTERM} and \texttt{SIGKILL} belong to the same
category: neither writes \texttt{down}. A stopped machine cannot wake
itself. Wake comes from \(\Omega.\operatorname{spawn}\), invoked by an
experimenter, a human, or the spawn verb of another machine (§4.3). The
E2 machine wakes with an unmatched lifecycle boundary and reprocesses
pending requests under the at-least-once semantics of transitions 6--7
in §3.3.1.

\hypertarget{liveness-boundary}{%
\subsection{Liveness Boundary}\label{liveness-boundary}}

Section 3.5 permits a damaged machine to lack a path back to
completeness when the repair organ is also damaged. The minimal
conditions for organ-level maintenance are: (i) the machine can run;
(ii) the receptionist of \texttt{c0} is alive or a successor has already
been prepared under the replacement discipline; (iii) an author is
available, either internal \texttt{c2} or one entering through a door;
and (iv) a description, protocol specification, or equivalent selection
criterion for the target member remains available. When all four hold, a
lost member can be rebuilt through the construction path. Section 4.5
directly witnesses this for \(C\) using a protocol specification
supplied by the experimenter. Without the fourth, an author may be
present but cannot know what to write. The first two conditions each
admit a liveness failure.

\textbf{The hand cannot reach.} Every morphology change passes through
the receptionist of \texttt{c0}. If it dies, requests cannot enter and a
new receptionist cannot be installed, because installing it would itself
require that receptionist. While alive, it can replace itself by adding
a successor, redirecting reception, and then retiring the old member, as
the transfer in E2-c demonstrates. Irrecoverability begins only when it
dies without a successor.

\textbf{The body cannot wake.} Suppose a message can kill \(R\) during
processing, whether by exhausting host resources or triggering a flaw in
a binding. The machine crashes and restarts by folding its ledger. The
message remains recorded and unfinished; it is redelivered and kills the
runtime again. Any recovery mechanism that redelivers an unfinished
event without isolating it enters the same loop: one bad point freezes
the machine, and the poison in history persists with history. This is
the other reason §3.3.4 rejects full replay. \(R\) guarantees only
truthful ledgers and foldable morphology (§3.3.2); identifying,
skipping, isolating, or backing off a poison message is engineering
policy. A poison message lies deeper than member failure because it
kills the physics rather than a member. Nor can a neighbor rescue this
case: detection \(\to\) spawn \(\to\) redelivery \(\to\) crash repeats,
defeating the recovery path of §4.3.

Both boundaries kill the individual. It stops in a well-defined but
irrecoverable state (§3.4.2), retaining identity and history. In the
poison-message case transitions continue---the machine alternates
between wake and crash---but no path returns it to completeness. \(G\)
does not die with the individual. It remains in its registry, in the
package of every descendant, and in \texttt{decl} on related machines.
If a lineage has survivors, reconstructing a machine is an ordinary
birth (§3.4.4). Section 4.5 establishes a stronger result: even when a
piece of \(G\) is missing, an author can re-create it from a
specification. The death of an individual is nevertheless real; once
\(H\) ends, it is not continued. Section 5.3 separates the accounting of
individual and lineage.

\hypertarget{discussion}{%
\section{Discussion}\label{discussion}}

Sections 2--4 describe one machine: the constructional core it inherits,
the substrate and obligations from which its architecture is derived,
and its behavior in operation. For an agent, the machine is not the end.
Three questions matter: what relation holds between agent and machine
(§5.1), what placing an agent inside a machine adds (§5.2), and what
this construction costs and does not promise (§5.3). The answers follow
from definitions in §3 and records in §4.

\hypertarget{the-relation-between-agent-and-machine}{%
\subsection{The Relation Between Agent and
Machine}\label{the-relation-between-agent-and-machine}}

The prevalent agent configuration is a model plus a periphery. The model
generates, while memory, tools, workflows, subagents, gateways, and
validators surround it (§1.1). Its lifecycle lies entirely in that
periphery: a loader installs it, a process manager restarts it, an image
copies it, a deployment script upgrades it, and a deployment location
supplies its identity. Whether the same model and prompt deployed twice
constitute one agent or two is decided by the deployer. The same
exterior decides whether a component written by the model at runtime
becomes part of the agent. In this configuration, the referent of
\emph{agent} is unspecified. Whether the model loop, tools, skills,
configuration, and deployment environment belong to the agent depends on
framework and deployment. The statement ``the agent manages and modifies
itself'' is therefore undecidable without first drawing the boundary
that determines whether the managed and modified objects count as the
agent.

Dalek instead has the configuration \textbf{machine plus members}.
Constitutive mechanisms dispersed through the periphery enter the
machine; remaining dependencies are listed in \(\Omega\) or enter as
inputs. Every lifecycle step is a machine transition. Its trigger and
material may come from outside, but whether a change counts as
installation, upgrade, reproduction, or continuation; what the machine
becomes afterward; how history continues; and what enters heredity are
all defined, relative to \(\Omega\), by \(G\), \(R\), and \(H\). The
model is one member \(L\): a program actor in \texttt{c2}, with an empty
binding, replaceable and able to coexist with another \(L\), whose loop
is text in \(G\) (§3.4.3).

\begin{longtable}[]{@{}
  >{\raggedright\arraybackslash}p{(\columnwidth - 4\tabcolsep) * \real{0.3333}}
  >{\raggedright\arraybackslash}p{(\columnwidth - 4\tabcolsep) * \real{0.3333}}
  >{\raggedright\arraybackslash}p{(\columnwidth - 4\tabcolsep) * \real{0.3333}}@{}}
\toprule\noalign{}
\begin{minipage}[b]{\linewidth}\raggedright
Operation
\end{minipage} & \begin{minipage}[b]{\linewidth}\raggedright
Model plus periphery
\end{minipage} & \begin{minipage}[b]{\linewidth}\raggedright
Machine plus members
\end{minipage} \\
\midrule\noalign{}
\endhead
\bottomrule\noalign{}
\endlastfoot
install a component & loader reads a file & \texttt{add} enters the
ledger and \texttt{placed} registers it in \(G\) (§§3.4.1--3.4.2) \\
restart & process manager restarts; state depends on peripheral
persistence & \(\Omega.\operatorname{spawn}\) starts \(R\); \(R\) folds
\(H\) and reconstructs morphology; \(\Sigma\) resets (§3.3.2) \\
copy & image plus deployment script & \texttt{decl\ →\ pack\ →\ spawn}
(§3.4.5) \\
upgrade & replace image or edit configuration & add a new member and
retire the old one; to change physics, produce an offspring (§§3.4.2,
3.3.5) \\
identity & deployment location or external name & \((G,H)\) (§3.4.6) \\
ownership of a model-written component & file system, loader, and
operator decide & it becomes a member through \texttt{c0} and is
inherited in \(G\) (§4.2) \\
position of the model & generative center, commonly called the agent &
one member: the author \\
maintainer & people and scripts in the periphery & repair paths and
authority belong to organs; targets, diagnoses, and triggers may enter
externally (§§4.3, 4.5) \\
\end{longtable}

Each cell in the left column follows a separate peripheral convention.
Each in the right is a transition defined in §3; an external event may
trigger it, but the machine defines its meaning. This is what §1.4 means
by moving the organizing center from task loop to lifecycle. The task
loop remains---it is the loop of \(L\)---but becomes an ordinary member
rather than the center of organization.

The subject of each \emph{self} is now fixed and appears at one of three
scales. At individual scale, self-maintenance preserves or restores the
relation between one machine's morphology, function, and capability and
its target \(Q\). At generational scale, self-reproduction relates
parent and offspring by copying \(G\), while self-evolution is change
along a lineage: candidate production, runnability testing, and
installation occur within one individual, but evolution completes only
when the change enters \(G\) and is inherited (§3.4.3). At population
scale, self-organization constructs the topology of doors among machines
(§§4.1, 4.3). The subject at no scale is \(L\). Its role is author: it
writes candidates and asks \texttt{c0} to install them through the same
route as an external requester (§3.4.3).

The model-plus-periphery configuration cannot state ``the agent manages
and modifies itself'' with a stable subject. The modifier is the model,
the modified objects surround the model, and maintenance and
adjudication remain external. Dalek's central contribution can be read
as constructing the subject of that sentence. Once constructed, the
sentence has an operational meaning: a member invokes \texttt{c0}
through \texttt{call} to alter \(M_t\) and \(G_t\) of this machine;
\(H\) of this machine witnesses the alteration; the portion entering
\(G\) is inherited by the lineage while \(H\) remains with the
individual; and organs of the same machine maintain it (§§4.3, 4.5).

``Unit of intelligence'' is a unit of individuation rather than a claim
about how intelligent the machine is. \(L\) supplies intelligence; the
machine supplies the unit in which intelligence is individualized,
maintained, and inherited. The medium is blind to the interior of \(L\),
which appears only through the ordinary actor and message interface. The
machine is fitted to the native properties of \(L\) (§2.2): built for
it, not centered on it.

\hypertarget{what-the-machine-adds}{%
\subsection{What the Machine Adds}\label{what-the-machine-adds}}

The difference made by the machine has two axes: computational
capability and constitution.

The capability axis concerns computational expressiveness. Exec
determines what can be computed and \(L\) what can be written; \(D\)
increases neither. With \(\Omega\) and \(L\) fixed, installing a new
member does not move the machine on this axis (§3.4.3). In E1, \(U\)
could write a file from the outset, and E1-a did so inside \(U\) without
adding anything to the machine (§4.2). No organ computes something
unavailable to \(U\).

The constitutive axis contains the entire difference. E1 adds a member
with a name and address (§3.2.2), complete text in \(H\) and
registration in \(G_t\) (§3.4.2), availability to every member,
reinstantiation after restart (§3.3.2), verbatim heredity during
reproduction (§3.4.5), and replacement by \texttt{add} plus
\texttt{retire} (§3.4.2). Machine capability grows by accumulation on
this axis.

Constitution is not mere wiring. Inventory-based systems connect parts,
and dynamically generated components may be called. The distinction is
the four obligations of §1.2. When a capability becomes a member of a
constructively defined machine, its construction language and admissible
transitions determine when it belongs to the machine, how it persists,
and who can admit it; the rules for producing legal successors are
inherited with them. For a one-off script, the distinction between the
two axes is nil. For a system of many parts that must be maintained,
copied, migrated, and upgraded, it is the distinction between having and
not having a system. ``It happened once'' is an event on the capability
axis. ``It remains defined''---membership, persistence, legal
transitions, and hereditary consequence always have a meaning---is a
property on the constitutive axis (§1.1).

Machine complexity grows in \(G\), not in \(R\). \(R\) recognizes only
addresses, \texttt{kind}, and \texttt{text} as a parameter (§3.3.3). A
machine may grow from three channels to arbitrarily many members and
organs without a dedicated branch in \(R\). This is the structure of a
universal constructor (§2.1): a more complex object lengthens its
description rather than its constructor. Three further facts about \(R\)
are independent. Renaming all of \(G\) leaves its behavior invariant, so
\(R\) is blind to organizational names (§3.3.3). \(R\) is text in
\texttt{G.world}, representable and replaceable across generations
(§3.3.5). The reproductive fixed point still holds after replacement
(§4.4).

The terms \emph{genotype} and \emph{phenotype} have precise referents
here. \(G_t\) is the heritable morphology folded by the registry, and
\(M_t\) the actual morphology folded by the medium (§3.4.2). A
synchronic comparison of \(\pi(M_t)\) and \(G_t\) identifies health when
they are isomorphic and damage when they diverge. A diachronic
comparison of \(G_{t+1}\) and \(G_t\) identifies mutation whenever they
differ, whether the change comes from registering a new member or from a
false registration. A mutation has passed selection when descendants
inherit it and still meet the acceptance condition. Once registration
completes, the new \(M_t\) and new \(G_t\) are again isomorphic.

\hypertarget{costs-and-limits}{%
\subsection{Costs and Limits}\label{costs-and-limits}}

A constructive definition makes both the cost and the capability
boundary precise. The costs follow from one set of medium choices:
ledgers append only, morphology is reconstructed from descriptions, and
history does not travel with descriptions.

\begin{enumerate}
\def\labelenumi{\arabic{enumi}.}
\tightlist
\item
  \textbf{No overwrite.} A \texttt{place} row cannot be rewritten and
  physical addresses are not reused. Modifying a member means adding a
  new one and retiring the old one. The new member receives a new
  physical address and may take over the same logical tag (§§3.4.2,
  4.4). Receptionists change by adding a new receptionist and retiring
  the old one (§4.3). \texttt{world} is fixed for an individual, so
  changing physics requires an offspring (§3.3.5). The model rejects
  overwriting an old member or old history, not online succession at one
  logical address.
\item
  \textbf{\(\Sigma\) does not survive restart.} A living machine is
  \((G,H,\Sigma)\); an individual is \((G,H)\). Restart reconstructs
  morphology rather than the volatile scene. A member that must survive
  reconstructs itself from \(H\) (§3.3.2).
\item
  \textbf{\(H\) is not inherited.} An offspring begins with an empty
  \(H\). Parental experience passes on only in forms admitted to \(G\),
  never as history (§3.4.6).
\item
  \textbf{Individuals die.} A receptionist may die without a successor,
  or a poison message may lock wake and crash into a loop, leaving an
  individual in a well-defined but irrecoverable state (§4.7). Survival
  of a lineage is not survival of an individual: when \(H\) ends, that
  individual ends. A lineage can produce another machine through
  ordinary birth; it does not recover this one.
\end{enumerate}

The capability boundary follows the same line. The machine supplies a
constitutive path and the semantics of persistence, recovery, and
heredity. It does not require intention itself to be internal. In the
experiments, what to change, what to want, and what counts as good are
supplied from outside the membrane.

\begin{enumerate}
\def\labelenumi{\arabic{enumi}.}
\tightlist
\item
  \textbf{The four selves are mechanism claims.} Tasks, specifications,
  and diagnoses determine what to change or repair and whether the
  result is wanted. The machine supplies how change and repair occur and
  how their products are inherited (§4.1). \emph{Self} attributes
  constitutive mechanisms and transition semantics to the machine; it
  does not imply autonomous intent or physical independence from
  \(\Omega\).
\item
  \textbf{Specifications are implicit in this machine.} \(G\) carries
  descriptions. The capabilities the machine ought to possess and their
  acceptance conditions reside in tasks and experimenter judgments and
  enter through doors (note 1 in §3.5). Self-maintenance is therefore
  stated relative to a fixed target \(Q\); autonomous production of that
  target is outside the paper (§3.4.2).
\item
  \textbf{Every live-model experiment is a single run.} The claim is
  that the path exists and that each step appears in a ledger, not a
  claim about success probability (§4).
\item
  \textbf{The membrane is a membrane of organization and ledger.} Direct
  file and network effects inside an invocation are neither recorded nor
  inherited. The claims concern constitutive actions (§§1.3, 3.2.3).
\end{enumerate}

The ledger is an instrument suited to these claims. By definition, every
constitutive action passes through \texttt{call} and enters a ledger
(§3.3.4). The statement ``the machine did \(x\)'' therefore refers to a
constitutive action that occurred if and only if its row exists. Whether
a property is witnessed is decided by applying the acceptance statement
of §4.1 to a ledger segment. What the instrument cannot
observe---invocation internals, host-side effects, and intention---is
precisely what the claims exclude. The scope of the instrument matches
the scope of the claims; this is why the paper uses ledgers rather than
self-report or demonstration as evidence.

\hypertarget{related-work}{%
\section{Related Work}\label{related-work}}

The object constructed here intersects several bodies of work that
usually select different units: an agent as a task loop, a managed
process, an improvement procedure, a running configuration, a computer
with \(L\) as processor, or a model that simulates itself. This paper
instead studies a lifecycle subject capable of bearing action,
constitutive change, history, and succession at once. We review those
bodies of work in turn and then state their relation to Dalek.

\hypertarget{agent-harnesses}{%
\subsection{Agent Harnesses}\label{agent-harnesses}}

A harness is a system around a foundation model. It orchestrates
execution and determines how the model reasons and plans, invokes tools
and acts, observes and manages context, stores artifacts, and evaluates
results \citep{Weng26}. OpenAI describes a harness as an environment
designed for a model: repository documentation, linters, and structural
tests enforced mechanically by continuous integration
\citep{Lopopolo26}. Anthropic organizes a planner, generator, and
evaluator as three agents; the evaluator operates a page through
Playwright, and one autonomous run may last hours \citep{Rajasekaran26}.
DeepSeek Harness makes every component, including the agent loop, a
plug-in. Its sessions are append-only event logs from which everything
visible to the model is derived. A dynamic package defined by the model
at runtime exists only in process memory and disappears on restart;
persistence across restarts follows a configuration path
\citep{deepseek2026harness}.

The harness itself has also become an object of modification.
Meta-Harness defines a harness as code that decides what to store,
retrieve, and expose to the model, then asks an outer coding agent to
search that code; each candidate is a file-system directory containing
source, score, and trace \citep{lee2026metaharness}. AHE gives each
component a file representation and attaches a falsifiable prediction to
every edit, while making the run directory, validator, and model
configuration read-only \citep{lin2026ahe}. Weng argues that evaluators
and permission controls should remain outside the evolutionary loop
\citep{Weng26}. Section 6.3 considers the improvement loop itself.

\hypertarget{systems}{%
\subsection{Systems}\label{systems}}

The account of machine state and admissible transition in §3 takes the
form of an operational semantics: configurations plus an inductively
defined transition relation \citep{Plo81}. THE arranges a system in six
layers and makes cooperation among sequential processes explicit through
synchronization statements, allowing logical correctness to be
established in advance and implementations to be tested exhaustively
\citep{Dijkstra68}. The Nucleus leaves policy outside the kernel, which
supplies only process creation, communication, and control; operating
systems and ordinary programs differ only in jurisdiction
\citep{BrinchHansen70}. The authors of UNIX observed that their system
had almost from the beginning been able to maintain itself \citep{RT78}.
The actor model presents members and effects uniformly as actors and
messages \citep{Hew73}. Agha specifies an actor's response to one
message as a finite set of outgoing messages, a finite set of newly
created actors, and a next behavior \citep{Agha85}. Lamport defines a
causal partial order through the happened-before relation without
relying on physical clocks \citep{Lam78}.

Recent agent runtimes bring the same concerns to agents. Agent libOS
gives an agent a process identity, process-local object memory, message
queue, tool table, syscall-mediated just-in-time tools, child processes,
capabilities, checkpoints, and durable recovery. Its central invariant
allows the action surface visible to the model to evolve without
implicitly expanding resource authority or permitted information flows
\citep{zhang2026agentlibos}. Shepherd makes execution traces
first-class: every action is a structured event, and a trace can be
inspected, forked, replayed, and rolled back \citep{yu2026shepherd}.
Cordis uses a calculus of dynamic composition for component loading and
unloading: each effect is handed to the runtime together with its
inverse, and declared dependencies drive activation and deactivation;
the implementation performs in-process replacement through hot module
replacement without restarting \citep{Cordis26}. Agent operating systems
place lifecycle mechanisms in a runtime that manages agents. Dalek
places the medium \(R\) and the organizing organs in \(G\), making them
heritable parts of the machine itself.

\hypertarget{recursive-self-improvement}{%
\subsection{Recursive
Self-Improvement}\label{recursive-self-improvement}}

Work on recursive self-improvement revolves around one loop: what is
changed, who evaluates it, and how a result is retained. STOP supplies a
minimal form. An improver optimizes an input program against a utility
function, then receives itself as input; model weights remain unchanged.
Mean performance rises across iterations under GPT-4 and falls under
weaker models \citep{zelikman2024stop}. The optimized object has
progressed from prompts through structured context, workflows, and
harness code to optimizer code \citep{Weng26}. DGM lets an agent modify
its own repository to produce descendants, selects parents from an
archive by performance and descendant count, and validates each change
on coding benchmarks \citep{zhang2025dgm}. Hyperagents combine a
meta-agent and task agent into one editable program, making the
modification procedure itself modifiable \citep{zhang2026hyperagents}.
At each level, the referent of ``who changes whom'' shifts. Meta\(^n\)
instead fixes a meta-operation, applies recursion only to its input, and
lets convergence determine depth \citep{kim2026metan}. Empirically,
under matched feedback and inference budgets, automatic harness
evolution does not consistently outperform simple test-time scaling and
shows limited held-out generalization \citep{wang2026rethinking}. Dalek
proposes no new improvement policy. It supplies constitutive semantics
for installing a candidate, witnessing what happened, forming a
successor, and carrying the result across generations.

\hypertarget{agents-and-von-neumann}{%
\subsection{Agents and von Neumann}\label{agents-and-von-neumann}}

Agent research has borrowed two ideas from von Neumann. The first is the
1945 stored-program computer: arithmetic, control, memory, input, and
output are separated, while instructions and numerical data reside
together in memory \(M\) and are selected and executed by the control
unit \citep{vN45}. This analogy casts \(L\) as a processor. The second
is the 1948 self-reproducing automaton: a description is separated from
the constructor that interprets it and the copier that duplicates it
\citep{vN48}. Recent work uses this construction to ask whether \(L\)
can simulate itself and treats such simulation as a threshold for
recursive improvement.

The first line begins with a correspondence of nouns and increasingly
imports mechanisms from computer architecture. L2MAC presents itself as
a stored-program computer: an instruction registry is program storage, a
file store is data storage, a control unit sequences instructions and
manages context, and \(L\) is the processor \citep{holt2024l2mac}. Mi et
al.~map a five-tuple of perception, cognition, memory, tools, and action
onto computer organs and transfer memory hierarchy, direct memory
access, and pipelining to agents \citep{mi2025building}. Lin et
al.~extend the mapping: \(L\) corresponds to a processor, KV cache to
cache, the context window to main memory, and an agent framework to an
operating system. Their architecture separates a probabilistic execution
plane that answers what can be computed from a deterministic control
plane that answers what should be computed \citep{lin2026modelnative}.

In the second line, Zhang et al.~connect recursive self-improvement to
the 1948 automaton. Constructor, copier, controller, and description
correspond to a universal simulator, replicator, supervisory program,
and joint description. A recursion theorem yields a program capable of
simulating itself and defines an introspection threshold
\citep{zhang2026selfreference}. The object is \(L\) itself. The paper
treats an LLM as a symbolic machine without a physical architecture,
making material self-reproduction irrelevant and replacing structural
self-description with simulation of functional dynamics. A complete
description of an LLM would contain all of its weights, yet a model
cannot access those weights during forward propagation; one proposed
alternative is an external approximate self-model. Scaffold code is one
of four levels of modification used to classify systems such as STOP and
DGM.

Earlier work on reflection asked how a description connects to a running
machine. RLL requires every system component to be represented in the
same formalism as ordinary knowledge so that modifying the description
modifies the system \citep{greiner1980rll}. Smith requires an embedded
account of the system and a causal connection between that account and
what it describes \citep{smith1984reflection}. AERA begins from a seed
and grows during operation, using addition and deletion instead of
whole-system replacement \citep{nivel2013bounded}. The Gödel Machine
encodes all initial code in an axiomatic system and permits a rewrite
only after proving that the rewrite increases utility; it asks when
modification is justified \citep{schmidhuber2003godel}.

On the side of individuality, autopoiesis defines a system whose own
operations produce its components and boundary \citep{MV80}. Among the
open problems McMullin identifies for the lineage of self-reproducing
automata are identity, bootstrapping, and actual growth of complexity
\citep{McM00}.

\hypertarget{relation-to-dalek}{%
\subsection{Relation to Dalek}\label{relation-to-dalek}}

The relation between Dalek and these works has three forms. Dalek
directly inherits two constructions from von Neumann: the 1945 treatment
of instructions as manipulable machine data \citep{vN45} and the 1948
separation of description, constructor, and copier \citep{vN48}. It also
inherits members and messages from the actor model
\citep{Hew73, Agha85}, the systems principle of leaving policy outside
the kernel \citep{BrinchHansen70}, and the state-transition form of
operational semantics \citep{Plo81}. The two von Neumann constructions
meet in one statement: \(G\) is a stored program in the 1945 sense whose
subject is the machine itself in the 1948 sense---organs, author, and
runtime---interpreted by \(A\) and copied by \(B\).

Other mechanisms converge independently. Meta-Harness records candidate
source, scores, and traces in file-system directories, while AHE
represents editable components as files and makes the run directory,
validator, and model configuration read-only
\citep{lee2026metaharness, lin2026ahe}. Agent libOS and Shepherd make
capabilities, checkpoints, and manipulable execution traces runtime
objects \citep{zhang2026agentlibos, yu2026shepherd}. These mechanisms
address the same kinds of problems as ledger \(H\), separation of author
and validator, bindings, and the medium boundary---facts that cannot be
recomputed, separation of authorship from judgment, and separation of
action surface from resource authority---but they are not the sources of
Dalek's construction.

The objects differ. Prior work respectively defines a task loop, a
managed process, an improvement procedure, a running configuration, a
computer with \(L\) as processor, or a model that simulates itself.
Dalek defines a \textbf{lifecycle subject}: an individual identified by
description and history, bounded relative to a host contract, whose
runtime specifies admissible constitutive transitions and whose rules
for producing a successor are inherited by that successor. Section 3
establishes that the four obligations of §1.2 hold jointly for this
object. The ledgers in §4 provide one constitutive path for each of
self-maintenance, self-evolution, self-reproduction, and
self-organization. Dalek does not separately invent messages, ledgers,
self-description, dynamic loading, compilation, or self-reproduction.
Its language model is an ordinary member, replaced, maintained, and
inherited through the same path as every other member.

\hypertarget{conclusion-and-future-work}{%
\section{Conclusion and Future Work}\label{conclusion-and-future-work}}

This paper gives a constructive definition of a class of agent machines
and supplies a running witness. Its purpose is not to strengthen one
task loop but to give continuously changing capabilities a subject that
can persist. Relative to an explicit host contract \(\Omega\), this
subject defines its members, boundary, history, and admissible
constitutive transitions through a finite medium. It describes itself
with \(G\) and distinguishes itself with \(H\). Its identity remains
decidable through installation, restart, and change, while an offspring
produced by reproduction is unambiguously distinct from its parent. A
large language model is no longer the organizing center surrounded by
peripheral structure; it is an author inside the machine that can itself
be replaced, maintained, and inherited.

The running records provide a constitutive path for maintenance,
evolution, reproduction, and organization by the machine itself. These
are not four capabilities attached to the machine but four directions of
one lifecycle across individual, generation, and population. The paper
establishes existence at the level of mechanism. Tasks, specifications,
and value judgments may still come from outside; whether and how a
change becomes part of the machine, persists, recovers, and is inherited
is now defined by the machine. Dalek is not a stronger agent loop. It is
a machine in which such loops can grow, change, and continue.

\hypertarget{future-work-conservative-extensions}{%
\subsection{Future Work: Conservative
Extensions}\label{future-work-conservative-extensions}}

The machine constructed here satisfies the four criteria of §1.2. Its
strict structure does not constrain future development; it gives each
development a defined path. New capabilities, organs, and protocols
enter the self-description as members through the same installation
path, are recorded in ledgers, and are inherited with the description.
The runtime need not grow with them; complexity grows in \(G\). Four
conservative extensions can enlarge this space while preserving the
architecture and criteria. They bring specification and selection into
the machine, bring \(L\) and \(U\) themselves into the construction
path, allow machines to form larger machines, and permit organized
capabilities to move and recombine across lineages.

\hypertarget{specifications-in-the-description}{%
\subsubsection{Specifications in the
Description}\label{specifications-in-the-description}}

The present machine can repair or modify itself against a given target
\(Q\), but \(Q\) comes from \(G\), prior history, or a specification
sent through the membrane (§§3.4.2 and note 1 in §3.5). A future \(G\)
could contain a specification distinct from its implementation
description, together with organs that hold, interpret, and compare
specifications. Implementation and specification would then be related
by an explicit satisfaction relation. A member holding a selection
criterion is structurally no different from one holding a specification;
whether a variation is worth retaining becomes internally decidable when
the specification carries an executable criterion. A specification would
remain quasi-quiescent text, copied and inherited with \(G\), while
ordinary members propose, modify, test, and adopt it through messages,
leaving evidence in \(H\).

This layer would allow self-maintenance to be stated against more than a
transient external target. A machine could preserve a specification
while changing implementations, recognize textually different members as
equivalent when both satisfy it, and allow contracts, acceptance
conditions, and governance rules to persist and be inherited with the
organization. ``Able to change itself'' would become ``able to change
itself under an explicit norm,'' giving long-term maintenance,
substitutable implementation, and auditable governance a common object.

\hypertarget{making-l-and-u-machines}{%
\subsubsection{\texorpdfstring{Making \(L\) and \(U\)
Machines}{Making L and U Machines}}\label{making-l-and-u-machines}}

\(L\) and \(U\) are already ordinary members of the medium, but their
interiors are supplied as black boxes. Each could be replaced by a
channel or Space preserving the same message protocol. \(U\) could
unfold into a seed compiler, compiler source, builder, tester, and
validator. \(L\) could unfold into a data pipeline, trainer, inference
engine, and evaluation loop, with training data and process as its
description and weights as the product of construction, just as an
executable process is produced from source. Once unfolded, exchanges
across channels would be requests and replies through doors, in the same
pattern by which \texttt{c0} obtains \texttt{decl} from \texttt{c1}
(§4.5). A facade member in the local channel would retain the session
and preserve the synchronous call boundary. Seed compilers,
accelerators, and foundation models could remain in \(\Omega\) or
external services; each machine could choose how far to unfold. The
exterior would still send text to \(L\) and candidates plus tests to
\(U\). This extension therefore connects directly to the scale protocol
in §7.1.3.

The resulting direction is maintenance and progressive internalization
of capability production. Compiler upgrades, model training, evaluation,
and rollback could be constructed, checked, replaced, recovered, and
inherited like other organs. A machine could gradually take
responsibility not only for existing capabilities but also for the
infrastructure that produces and judges them. Self-sufficiency would
cease to be a binary choice between perpetual dependence on an external
service and rebuilding an entire technical stack at birth.

\hypertarget{machine-as-actor}{%
\subsubsection{Machine as Actor}\label{machine-as-actor}}

A pair of dual organizational protocols could connect scales. An
\textbf{encapsulation protocol} would present a channel driven by its
Space, or an entire Space carrying its own \(G\) and \(H\), as one
member of an outer machine through a bridge and receptionist. A
\textbf{refinement protocol} would replace one member with a channel or
Space implementing the same message protocol. Each protocol must state
which observations replacement preserves---replies, asynchronous
messages, faults, lifecycle behavior, and resource effects---but need
not require identical structures at both levels, nor require refinement
after encapsulation to reconstruct the original. A bridge remains an
ordinary member or door; every cross-boundary effect remains a message;
the inner machine retains its own boundary, identity, and history. The
central research problem is observational equivalence and asynchronous
control flow across scale. A call returns synchronously while a door
returns empty; how a machine presented as a member replies to a call is
the first question the protocol must answer.

The value is organizational growth without a new meta-framework at every
scale. A complex organ may acquire an independent maintenance and fault
boundary, while a complete machine may become an organ of a larger
machine. Every level uses the same organizational primitives. Because
humans and agents alike enter through actors and messages, role
assignment, supervision, decision, and succession can become ordinary
organs in channels; governance itself can recursively enter the
organization it governs. Complexity grows in \(G\) without forcing \(R\)
to grow with organizational depth.

\hypertarget{sharing-and-propagating-fragments-of-capability}{%
\subsubsection{Sharing and Propagating Fragments of
Capability}\label{sharing-and-propagating-fragments-of-capability}}

Alongside a complete \(G\), the system could define a portable
\textbf{\(G\) fragment} describing one actor, one channel, or a set of
channels with internal topology, together with explicit imports,
exports, interface versions, and \texttt{world} requirements. Export and
import organs would extract, inspect, rename, bind dependencies, and
reauthorize the fragment. Once accepted, \texttt{c0} would construct it
through the existing path, \(H\) would record it, and \texttt{c1} would
register it. The protocol would transmit a quasi-quiescent construction
description rather than the original instance's \(H\), volatile state,
live endpoints, bindings, or host-side effects. The receiver would
revalidate and reauthorize the fragment inside its own boundary.

This is more than another form of code distribution. The unit of reuse
could be a compiler and test organ, a research or production pipeline,
or a group of channels carrying internal governance. An organization
developed by one lineage could be accepted by another and recombined
with a fragment from a third. A capability would move from a trick
inside one session to an inspectable, maintainable, composable, and
heritable organization.

\appendix

\hypertarget{notation}{%
\section*{Notation}\label{notation}}
\addcontentsline{toc}{section}{Notation}

\textbf{Individual \(=(G,H)\).} \(G\) answers ``what am I?'' and \(H\)
answers ``which one am I?'' (§3.4.6). A running machine is
\((G,H,\Sigma)\); \(\Sigma\) does not enter the ledger and resets on
restart (§3.3.2). ``One machine'' means one Space, driven by the \(R\)
it carries and runnable on any host satisfying \(\Omega\) (§3.2.1). The
letter \(E\) is used only for the 1948 constructional model
\(E=(A+B+C+D)+G\) in §2; E1--E5 in §4 are experiment identifiers and
denote no machine.

Roles, organs, and medium:

\begin{longtable}[]{@{}
  >{\raggedright\arraybackslash}p{(\columnwidth - 4\tabcolsep) * \real{0.3333}}
  >{\raggedright\arraybackslash}p{(\columnwidth - 4\tabcolsep) * \real{0.3333}}
  >{\raggedright\arraybackslash}p{(\columnwidth - 4\tabcolsep) * \real{0.3333}}@{}}
\toprule\noalign{}
\begin{minipage}[b]{\linewidth}\raggedright
Symbol
\end{minipage} & \begin{minipage}[b]{\linewidth}\raggedright
Name
\end{minipage} & \begin{minipage}[b]{\linewidth}\raggedright
Definition
\end{minipage} \\
\midrule\noalign{}
\endhead
\bottomrule\noalign{}
\endlastfoot
\(\Omega\) & host contract & execution, storage, and networking;
supplied at the destination and does not travel with the machine
(§3.1) \\
\(R\) & runtime & the structural fixed point: instantiates actors,
delivers messages, folds ledgers, and executes syscalls; blind to
function and organization; inherited across a lineage in
\texttt{G.world} (§3.3) \\
\(G\) & self-description & the complete inventory of machine morphology,
including \texttt{world}, construction organs, and the agent loop;
answers ``what am I?''; modifiable at runtime and inherited through
reproduction (§§3.4.2, 3.4.6) \\
\(H\) & ledger & complete append-only, medium-stamped operational
history; answers ``which one am I?''; not inherited (§§3.2.3, 3.4.6) \\
\(A\) & constructor & the sole interpreter of \(G\): \texttt{realize}
(§3.4.1) \\
\(B\) & copier & copies without reading: \texttt{pack} (§§3.4.1,
3.4.5) \\
\(C\) & controller & sequences
\texttt{decl\ →\ pack\ →\ spawn\ →\ build\ →\ start} (§3.4.4) \\
\(D\) & general capability producer & \(L+U\); can produce members,
loops, and the runtime of an offspring (§3.4.3) \\
\(L\) & large language model, the author & generates capability
candidates; an ordinary unprivileged, replaceable member (§§2.2,
3.4.3) \\
\(U\) & compiler & turns candidate text into an installable part,
separate from the author (§3.4.3) \\
\texttt{c0} & construction organ & carries \(A\), \(B\), and \(C\); the
only channel that interprets \(G\) (§3.4.1) \\
\texttt{c1} & registry & residence of \(G\);
\(\texttt{decl}=G_0\oplus\texttt{placed}\ominus\texttt{retired}\)
(§3.4.2) \\
\texttt{c2} & author organ & carries \(D=L+U\); \texttt{bind={[}{]}};
changes morphology by the same route as an external requester
(§3.4.3) \\
\end{longtable}

Morphology, targets, and derived quantities:

\begin{longtable}[]{@{}
  >{\raggedright\arraybackslash}p{(\columnwidth - 4\tabcolsep) * \real{0.3333}}
  >{\raggedright\arraybackslash}p{(\columnwidth - 4\tabcolsep) * \real{0.3333}}
  >{\raggedright\arraybackslash}p{(\columnwidth - 4\tabcolsep) * \real{0.3333}}@{}}
\toprule\noalign{}
\begin{minipage}[b]{\linewidth}\raggedright
Symbol
\end{minipage} & \begin{minipage}[b]{\linewidth}\raggedright
Name
\end{minipage} & \begin{minipage}[b]{\linewidth}\raggedright
Definition
\end{minipage} \\
\midrule\noalign{}
\endhead
\bottomrule\noalign{}
\endlastfoot
\(G_t\) & heritable morphology & fold by \texttt{c1} over its ledger:
the structure the machine declares should be inherited and regenerated
(§3.4.2) \\
\(M_t\) & actual morphology & fold by \(R\) over every ledger: the
structure currently alive (§3.4.2) \\
\(\pi\) & provenance projection & retains everything installed by the
constructor and \texttt{c0} members installed through the root door;
removes birth certificates, temporary doors, and retired members; at
rest \texttt{c0} maintains \(\pi(M_t)\cong G_t\) (§3.4.2) \\
\(Q\) & maintenance target & fixed before repair: \(G_t\) for
morphology, a member specification for function, or runnability and
reproduction for capability; supplied by \(G\), prior history, or an
external specification (§§3.4.2, 4.1) \\
\texttt{world} & physics & root field of \(G\) containing
\(\omega\)-bind, loader, and \(R\); fixed for an individual and variable
across a lineage (§3.3.5) \\
\(\Sigma\) & volatile state & internal member state; absent from the
ledger and reset by restart (§3.3.2) \\
\(P\) & packaged form & \texttt{P\ =\ pack(G)}: three \texttt{world}
files plus \texttt{G.json}; dead, portable across hosts, and comparable
(§§3.4.1, 3.4.5, Appendix A.6) \\
\texttt{decl} & obtain description & operation by which \texttt{c1}
returns a snapshot of \(G_t\); fixed throughout \texttt{pack},
\texttt{spawn}, and \texttt{start} (§§3.4.2, 3.4.4) \\
\texttt{realize\ /\ pack} & construct / copy & \texttt{realize} reads
structure and transports text (\(A\)); \texttt{pack} copies all of \(G\)
without reading it (\(B\)) (§3.4.1) \\
\end{longtable}

Vocabulary: actor, message, channel, and Space are the basic
construction units and three membranes (§3.2.1). A \textbf{member} is an
actor registered in a channel. Doors and the root door are defined in
§§3.2.4 and 3.4.4. \textbf{Syscalls} are the three morphology-writing
calls \texttt{channel.create}, \texttt{channel.add.actor}, and
\texttt{channel.retire.actor}; \texttt{spawn} and \texttt{stop} are
world verbs (§3.2.2 and Appendix A.2). \textbf{Outside the membrane}
means anything beyond the machine boundary.

\hypertarget{transition-table-and-abi}{%
\section{Transition Table and ABI}\label{transition-table-and-abi}}

\hypertarget{state-and-event}{%
\subsection{State and Event}\label{state-and-event}}

State consists of one append-only ledger per channel and one cursor per
actor. An event is a \texttt{msg} row addressed to some address and
carrying no \texttt{run} marker. One event invokes that member once, run
to completion. Nested calls form the call stack; only events advance a
cursor.

\hypertarget{complete-transition-table}{%
\subsection{Complete Transition Table}\label{complete-transition-table}}

\begin{longtable}[]{@{}
  >{\raggedright\arraybackslash}p{(\columnwidth - 4\tabcolsep) * \real{0.3333}}
  >{\raggedright\arraybackslash}p{(\columnwidth - 4\tabcolsep) * \real{0.3333}}
  >{\raggedright\arraybackslash}p{(\columnwidth - 4\tabcolsep) * \real{0.3333}}@{}}
\toprule\noalign{}
\begin{minipage}[b]{\linewidth}\raggedright
\texttt{kind}
\end{minipage} & \begin{minipage}[b]{\linewidth}\raggedright
Instantiation, once when \texttt{place} is folded
\end{minipage} & \begin{minipage}[b]{\linewidth}\raggedright
For each message
\end{minipage} \\
\midrule\noalign{}
\endhead
\bottomrule\noalign{}
\endlastfoot
program & \texttt{Exec.load(text,\ \{call,\ me,\ channel\})}: execute
source once; it must define \texttt{run(m)}, where
\texttt{m\ =\ \{seq,\ from,\ to,\ body,\ channel\}} & \texttt{run(m)};
the return value is the reply; \texttt{None} or empty means no reply \\
door & none; \texttt{text} is an endpoint address & deliver unchanged
through \texttt{Port.send}, stamp the current channel endpoint, return
empty \\
install actor (\texttt{syscall\ channel.add.actor}) & \(R\) allocates a
unique tag in the live routing table, adding a numeric suffix on
collision, and writes a \texttt{place} row containing the
\textbf{complete text} & receipt \texttt{channel/tag} \\
\end{longtable}

The other two syscalls are \texttt{channel.create}, which creates a
channel, and \texttt{channel.retire.actor}, which appends a
\texttt{retire} row. Addresses are never reused and the current
receptionist cannot retire. The world verbs are \texttt{spawn}, which
starts an offspring process, and \texttt{stop}, which performs legal
shutdown.

\hypertarget{address-space-of-call}{%
\subsection{\texorpdfstring{Address Space of
\texttt{call}}{Address Space of call}}\label{address-space-of-call}}

\begin{longtable}[]{@{}
  >{\raggedright\arraybackslash}p{(\columnwidth - 4\tabcolsep) * \real{0.3333}}
  >{\raggedright\arraybackslash}p{(\columnwidth - 4\tabcolsep) * \real{0.3333}}
  >{\raggedright\arraybackslash}p{(\columnwidth - 4\tabcolsep) * \real{0.3333}}@{}}
\toprule\noalign{}
\begin{minipage}[b]{\linewidth}\raggedright
Address
\end{minipage} & \begin{minipage}[b]{\linewidth}\raggedright
Semantics
\end{minipage} & \begin{minipage}[b]{\linewidth}\raggedright
Required binding
\end{minipage} \\
\midrule\noalign{}
\endhead
\bottomrule\noalign{}
\endlastfoot
tag & member in the current channel; resolved at call time & none \\
\texttt{0} & read medium: \texttt{show\ {[}a{]}\ {[}b{]}} returns ledger
rows; \texttt{who} returns the current member table. Read is open to
every member and records only a fact row. & none \\
door tag & send outside the membrane and return empty & none \\
\texttt{channel.create}, \texttt{channel.add.actor},
\texttt{channel.retire.actor} & write morphology & \texttt{syscall} \\
\texttt{spawn}, \texttt{stop} & act on the world & binding of the same
name \\
\end{longtable}

A call to a nonexistent address is discarded. A call to a member invokes
it synchronously, and the member's reply is the return value.

\hypertarget{ledger-rows}{%
\subsection{Ledger Rows}\label{ledger-rows}}

Each channel has one JSON Lines file written by one writer, the local
\(R\):

\begin{longtable}[]{@{}
  >{\raggedright\arraybackslash}p{(\columnwidth - 4\tabcolsep) * \real{0.3333}}
  >{\raggedright\arraybackslash}p{(\columnwidth - 4\tabcolsep) * \real{0.3333}}
  >{\raggedright\arraybackslash}p{(\columnwidth - 4\tabcolsep) * \real{0.3333}}@{}}
\toprule\noalign{}
\begin{minipage}[b]{\linewidth}\raggedright
\texttt{k}
\end{minipage} & \begin{minipage}[b]{\linewidth}\raggedright
Principal fields
\end{minipage} & \begin{minipage}[b]{\linewidth}\raggedright
Writer
\end{minipage} \\
\midrule\noalign{}
\endhead
\bottomrule\noalign{}
\endlastfoot
\texttt{place} &
\texttt{seq,\ addr,\ kind,\ text,\ bind,\ in,\ by,\ tag,\ iface?} &
\(R\) executing \texttt{add}; contains complete \texttt{text} \\
\texttt{retire} & \texttt{seq,\ addr} & \(R\) executing
\texttt{retire} \\
\texttt{msg} & \texttt{seq,\ from,\ to,\ body,\ run?,\ at?,\ by?} &
requests and replies, messages copied from doors, syscall receipts, and
fact rows from \texttt{0}; \texttt{run} is the containing event's
sequence; \texttt{at} is inbox offset \\
\texttt{step} & \texttt{seq,\ actor,\ upto,\ out,\ err?,\ run?} & closes
an invocation with its cause, every emitted frame, and any exception \\
\end{longtable}

The medium stamps \texttt{from}. An external arrival is signed by its
corresponding door, or \texttt{door} if no corresponding door exists. A
numeric physical address \texttt{addr} is a position in \(H\): it only
increases, is never reused, and is not inherited. The logical address
\texttt{channel/tag} is written into \(G\) and inherited.

\hypertarget{root-door-and-inboxes}{%
\subsection{Root Door and Inboxes}\label{root-door-and-inboxes}}

\begin{itemize}
\tightlist
\item
  Every channel has an inbox, the receiving side of Port. An arrival
  becomes a \texttt{msg} for the member whose \texttt{place} row carries
  the \texttt{in} mark.
\item
  The root door belongs to \(R\) at Space level, exists before every
  channel, and is absent from \(G\). It is \textbf{open iff no ledger
  contains a \texttt{msg} row}, a property derived from ledgers rather
  than hidden state. While open, it accepts \texttt{channel.create},
  \texttt{channel.add.actor}---recorded with \texttt{by=\_root}---and
  the first message, \texttt{start} with \(G\) in its body. That message
  also closes the door. Every later root-door arrival is ignored.
\item
  Lifecycle boundaries appear only in the root channel. Birth is
  \texttt{start}; each subsequent wake and legal shutdown is an
  \texttt{up} or \texttt{down} row from \texttt{\_root} to the
  receptionist. Reinstantiation of members is derived from that
  machine-level event and each member's \texttt{place/retire} history.
\end{itemize}

\hypertarget{packaged-form-p-and-startup}{%
\subsection{\texorpdfstring{Packaged Form \(P\) and
Startup}{Packaged Form P and Startup}}\label{packaged-form-p-and-startup}}

\begin{Shaded}
\begin{Highlighting}[]
\NormalTok{P/omega.py runtime.py init.py   three world files, byte{-}identical to G.world}
\NormalTok{P/G.json                        description unchanged, produced by B}
\NormalTok{P/h/\textless{}channel\textgreater{}.jsonl, h/\_order   ledgers and medium boot index, created at runtime}
\NormalTok{P/in/\textless{}box\textgreater{}.jsonl                inboxes for \_root and every channel, created at runtime}
\NormalTok{P/spawn/\textless{}name\textgreater{}/                 offspring P, created during reproduction; lineage is path}
\NormalTok{python init.py \textless{}P\textgreater{} [{-}{-}serve]    start R, fold existing ledgers, drive; do not read G}
\end{Highlighting}
\end{Shaded}

\hypertarget{evidence-index}{%
\section{Evidence Index}\label{evidence-index}}

The following files contain the original rows cited by the tables in §4.
The ledgers are released with the source, and snapshots are included
under \texttt{evidence/runs/}; file names in the table are relative to
that directory unless another location is shown. Credentials have been
redacted (\texttt{KEY} in the source of \(L\) is \texttt{sk-***}).
Original records are not rewritten. E1 and E2 predate tag addressing, so
external receipts use numeric addresses such as \texttt{c2/5}. E2
predates tightened lifecycle semantics, so restart events were injected
into each channel.

\begin{longtable}[]{@{}
  >{\raggedright\arraybackslash}p{(\columnwidth - 6\tabcolsep) * \real{0.2500}}
  >{\raggedright\arraybackslash}p{(\columnwidth - 6\tabcolsep) * \real{0.2500}}
  >{\raggedright\arraybackslash}p{(\columnwidth - 6\tabcolsep) * \real{0.2500}}
  >{\raggedright\arraybackslash}p{(\columnwidth - 6\tabcolsep) * \real{0.2500}}@{}}
\toprule\noalign{}
\begin{minipage}[b]{\linewidth}\raggedright
Experiment
\end{minipage} & \begin{minipage}[b]{\linewidth}\raggedright
Author
\end{minipage} & \begin{minipage}[b]{\linewidth}\raggedright
Evidence file
\end{minipage} & \begin{minipage}[b]{\linewidth}\raggedright
Section
\end{minipage} \\
\midrule\noalign{}
\endhead
\bottomrule\noalign{}
\endlastfoot
E1-b parent & \texttt{deepseek-chat} & \texttt{task0-deepseek-c2.jsonl}
& §4.2 \\
E1-b offspring & --- & \texttt{task0-deepseek-child-c2.jsonl} & §4.2 \\
E2-b \texttt{d0} installation & \texttt{deepseek-v4-pro} &
\texttt{task1-d0-c2.jsonl}; \texttt{c3} and \texttt{c4}:
\texttt{task1-d0-c3.jsonl}, \texttt{task1-d0-c4.jsonl} & §4.3 \\
E2-b/E2-c \texttt{d1} & same & \texttt{task1-d1-c4.jsonl} for
self-organization, defect, and wake; \texttt{task1-d1-c2.jsonl} for
repair & §4.3 \\
E2-b \texttt{d2} & --- & \texttt{task1-d2-c4.jsonl} & §4.3 \\
E3 & stub, mechanism layer & \texttt{t/test\_c0.py}, E3 driver: local
damage, shutdown/wake, hard kill & §4.6 \\
E4 & stub, mechanism layer & \texttt{t/test\_c0.py}, E4 driver & §4.4 \\
E5-b parent & \texttt{deepseek-v4-pro} &
\texttt{restore-deepseek-c0.jsonl}, \texttt{-c1.jsonl},
\texttt{-c2.jsonl} & §4.5 \\
E5-b offspring & --- & \texttt{restore-deepseek-kid-c0.jsonl},
\texttt{-kid-c1.jsonl} & §4.5 \\
E5-b grandchild & --- & \texttt{restore-deepseek-grand-c0.jsonl} &
§4.5 \\
\end{longtable}

  \bibliography{refs/references.bib}

\end{document}